\documentclass[10pt]{article}
\usepackage[margin=1in]{geometry}
\usepackage{authblk}

\usepackage[]{graphicx} 
\usepackage[natbib]{biblatex}
\usepackage{color} 
\usepackage{eufrak}
\usepackage{hyperref}
\usepackage{bm}
\usepackage{amsfonts,amsmath,amssymb}

\newcommand{\tr}{\ensuremath\mathfrak{tr}}
\newcommand{\Cfull}{\ensuremath C_\mathrm{full}}
\newcommand{\Cdiag}{\ensuremath C_\mathrm{diag}}
\newcommand{\R}{\ensuremath \mathbb{R}}

\title{Neural Variability Enhances Artificial Network Robustness}
\date{June 11, 2026}
\author[1]{Robin Preble}
\author[2]{Praveen Venkatesh}
\author[2]{Stefan Mihalas}
\author[1,*]{Kameron Decker Harris}
\affil[1]{Department of Computer Science, Western Washington University, Bellingham, WA 98225}
\affil[2]{Allen Institute, Seattle, WA 98195}
\affil[*]{Corresponding author: {\tt harri267@wwu.edu}}

\begin{document}

\maketitle

\begin{abstract}
Neural responses in cortex exhibit substantial trial-to-trial variability in response to repeated stimuli, while peripheral sensory neurons respond far more consistently, leading many to wonder whether stochasticity may carry meaning. 
Existing work has argued that noise and signal correlations may be optimized for discrimination in animals, whereas artificial neural network (ANN) studies have shown similar benefits of noise in machine learning tasks, although most ANN work has neglected the effects of correlations.
Here we investigate whether correlated noise improves the robustness of artificial neural networks to adversarial attacks and naturalistic image modifications. 
Using the covariance of activations under modified versus clean inputs, we find that structured noise may significantly improve network robustness.
Robustness to naturalistic image modifications benefits most from structure, but this structure transfers poorly across modification types.
In contrast, noise structure from adversarial attacks can generalize to other kinds of attacks.
These results suggest that structured noise in ANN activations generally improves robustness, establishing a biologically plausible strategy for creating robust artificial neural networks that only relies on local information. 
\end{abstract}

\section{Introduction}

The brain is noisy, but whether this is beneficial or captures Bayesian notions of uncertainty is a matter of debate  \citep{stochastic_resonance,bayesian_inference,sampling_based}. 
At the same time, stochastic artificial neural networks (ANNs) have been explored for tractable theory \citep{chizat_lazy_2018,jacot_neural_2018}, Bayesian inference \citep{neal_bayesian_1996}, and improved generalization \citep{srivastava_dropout_nodate} or robustness \citep{zantedeschi_efficient_2017}.

ANNs are sensitive to perturbations that would not fool an animal.
These can be crafted with or without knowledge of the model weights, known as white or black box attacks.
\cite{zantedeschi_efficient_2017} proposed adding Gaussian noise to network inputs to defend against these attacks, and this can be certified \citep{cohen_certified_2019}.
Parametric noise injection (PNI) adds noise to either activations or weights with diagonal noise covariance \citep{parametric_noise}, while
colored noise injection (CNI) allows for covariance structure (low-rank plus diagonal) learned via backpropagation and was applied to the weights \citep{zheltonozhskii_colored_2020}.
\cite{vonenets} showed that unlearned V1-like features combined with stochastic spiking also protected in image classification tasks, which could be explained by the interaction of representation geometry and noise \citep{dapello_neural_2021}.

Correlations have been used to explain diverse phenomena in neuroscience \citep{de_la_rocha_correlation_2007}, and there are even models of optimal correlations, e.g.\ the {\em sign rule} which states that optimal noise and signal correlations have opposite signs \citep{hu_sign_2014}.
Motivated by recordings of noisy activity in mouse V1, \cite{venkatesh_role_2024} hypothesized that optimal noise would have a covariance that shrinks perpendicular to the optimal decision boundary (as in the sign rule), with the greatest variance in task-irrelevant directions. 
This hypothesis was supported by analysis of mouse V1 data, and they implemented a neural network with a noisy layer and injected noise with covariance structure defined by rotations of some images and observed that it provides robustness to rotations of other images, which we expand upon here.
We were motivated by this work and its neural relevance to test structured noise derived from other kinds of modifications.

\begin{figure*}[t]
\centering
\includegraphics[width=0.8\linewidth]{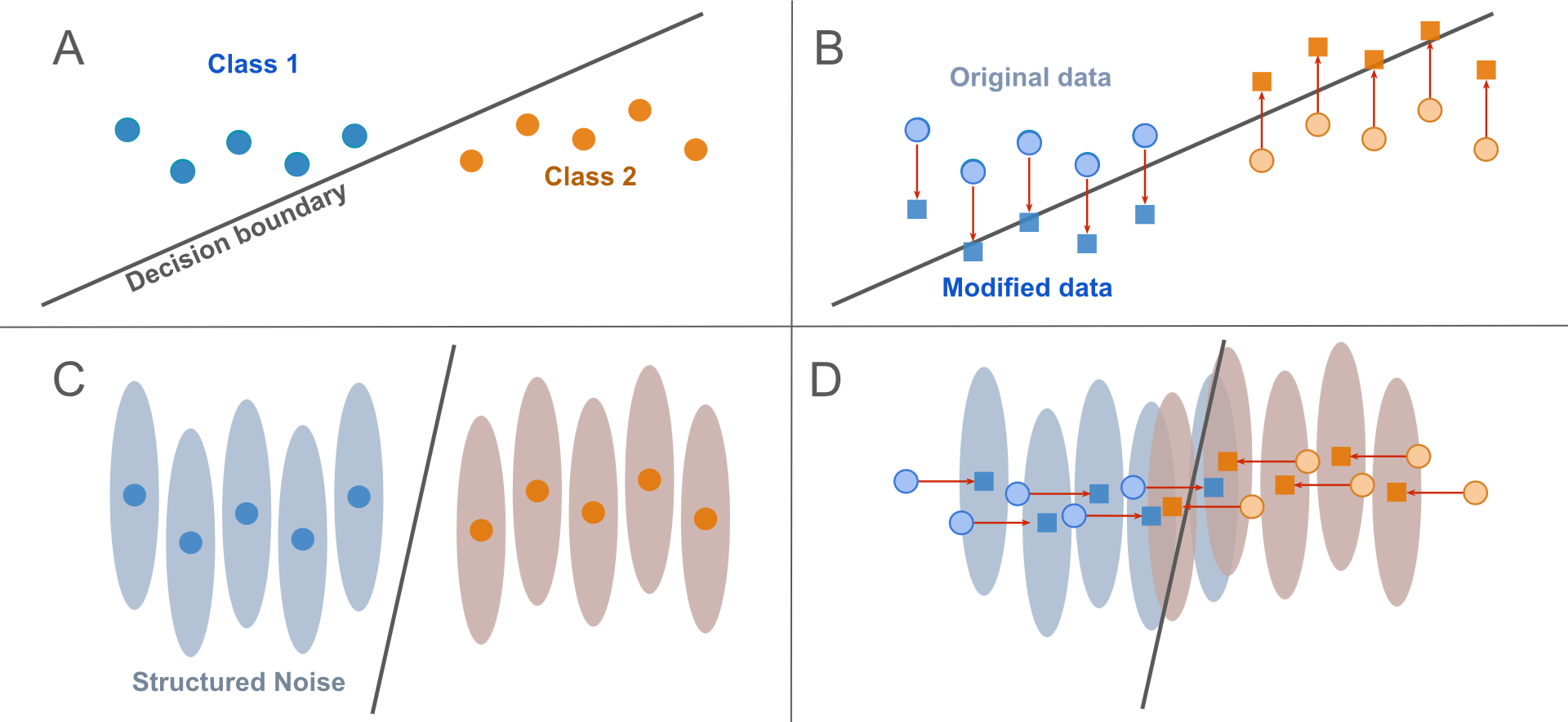}
\caption{\textbf{Noise structure can improve robustness.} 
A) Representation of two classes separated by a learned decision boundary. 
The vertical dimension isn't relevant to the task, but due to limited data the decision boundary varies over this dimension.
B) The classifier perfectly separates clean training data but is not robust to modifications such as adversarial attacks that move samples across the boundary. 
C) Structured noise (multivariate Gaussian, depicted as ellipses) is fit to the adversarial perturbations, and the classifier is retrained on the noisy data.
Noiseless representations are drawn in the center of the ellipses.
D) The retrained decision boundary has a larger margin and robustness against further modifications.
Unstructured noise, on the other hand, would be circular and could result in a smaller margin and worse overlap of the two classes.
}
\label{fig:diagram}
\end{figure*}

In this paper, we ask whether injecting structured noise into a neural network yields greater robustness than unstructured noise.
We hypothesize that noise where the covariance structure is derived directly from intermediate layer activations, rather than being learned (as in PNI/CNI), may improve robustness (Fig.~\ref{fig:diagram}).
If structured noise improves robustness in ANNs, this lends additional strength to the idea that the structured noise observed in the brain serves a purpose. 
As in \cite{venkatesh_role_2024}, we use Gaussian noise with a covariance calculated from the difference in model activations at a given layer with clean vs.\ modified inputs. 
We selected Gaussian noise because it is a simple model of noise with correlations.
In our approach, noise has the greatest variability in the directions that are most affected by the given modification. 
We chose this technique because increasing the variance in directions that the model should be invariant to pads the margin, while lowering variance in directions relevant for classification avoids overlap between representations of different classes.
Training on clean data with these noisy representations adjusts the decision boundary to be invariant to class-irrelevant data, without creating confusion between classes.
Our method depends on only layer-specific activations and is biologically plausible, since Hebbian mechanisms may shape noise covariance \citep{scott_beyond_2021,eppler_representational_2026}.

\section{Methods}

\subsection{Networks and Input Modifications}

Here we detail the architecture of our networks and the ways we attack or perturb the input, generally referred to as {\em modifications}.

\subsubsection{Architecture} 
Our base model was a standard neural network, based on the classic LeNet architecture \citep{lecun_gradient-based_1998}, with 3 convolutional layers interleaved with max-pooling followed by 3 fully-connected layers.
We trained on the Fashion MNIST \citep[FMNIST;][]{xiao_fashion-mnist_2017} dataset for 10 epochs using $\tanh$ activations between all layers excluding the last, Adam optimizer with learning rate 0.001, cross entropy loss, and batch size 64. 
Layer $\ell$ activations are denoted $x_\ell \in \R^{n_\ell}$ for $\ell = 0, \ldots, 6$ with $x_0$ for the input and $x_6$ the classifier logits.
In the appendix, we present similar results with a vision transformer (ViT) and CIFAR-10.

\subsubsection{Adversarial Attacks}
To investigate the model’s robustness, we compared its performance on data that has undergone a variety of modifications, including adversarially attacked data.
We used the Adversarial Robustness Toolbox \citep[ART;][]{art} to generate attacks and compared a range of attack strengths $\varepsilon = 0.001$ to 0.2 for each experiment. 
See Table \ref{tab:mod_descriptions} in the appendix for a list of attacks. 
To compare our results to established methods for defending against adversarial attacks, we used ART to implement Gaussian Data Augmentation \citep{zantedeschi_efficient_2017} and Adversarial Training \citep{tramer2020ensembleadversarialtrainingattacks}. Specifically, we used Gaussian augmentation on 100\% of the data during both training and evaluation and adversarial training with Projected Gradient Descent (PGD). 

\subsubsection{Naturalistic Modifications}
We used the both the imagecorruptions package \citep{michaelis2019dragon} and torchvision \citep{marcel_torchvision_2010,torchvision2016} to create modified images using a variety of perturbations shown in Tab.~\ref{tab:mod_descriptions}. 
Gaussian blur was excluded since we found it did not significantly affect the models. 

For image modifications from the imagecorruptions package, corruption severity is set by an integer between 1 and 5. 
This library requires that images are at least $32 \times 32$ with values ranging from 0 to 255, whereas FMNIST is $28 \times 28$ with values ranging from $[0,1]$.
To solve these issues, we padded images with zeros to make them $32 \times 32$ before processing and converted them to an unsigned byte format.
After processing, we cropped them to $28 \times 28$, converted them to a float format, converting them back to grayscale if necessary and clamping values to a range from $[0,1]$.
For torchvision modifications, we selected a base value for the transformation strength and multiplied it by a scaling factor between $0.2$ and $2.0$. 
We also developed random obstruction: a black square is placed over a random part of the image, with its side length defined as $0.4 s H $, where $s$ is a scaling factor in the range $0.2 \leq s \leq 2.0$ and $H$ is height of the (square) image.

\begin{figure*}[ht]
\centering
\includegraphics[width=1.0\linewidth]{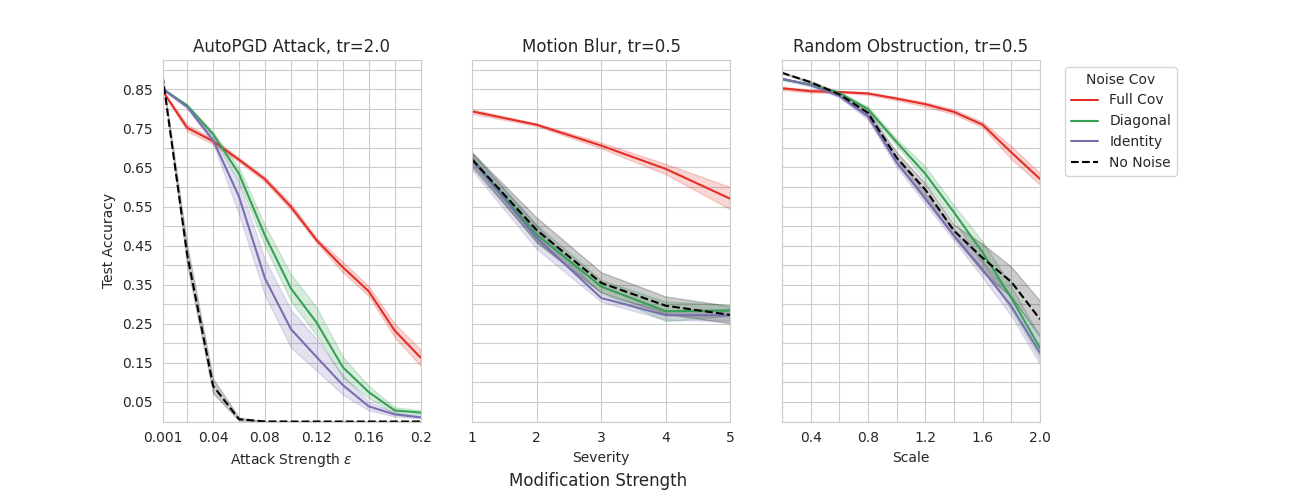}
\caption{\textbf{Structured noise improves robustness.}
Noise with a covariance derived from the base model's responses to modified data is injected into the activations of the second convolutional layer ($L=2$). 
Noise covariance settings include full covariance, diagonal covariance, identity covariance, and no noise.
(Left) Mean test accuracy against AutoPGD attack plotted against attack strength.
(Center) Mean test accuracy against range of motion blur severity.
(Right) Mean test accuracy against random obstruction, where a black square of a given size is placed in a random location.
}
\label{fig:ItWorks}
\end{figure*}

\subsection{Noisy Layer Models}

We inject noise into the activations $x_L$ of a selected noisy layer $L$ of the network.
This noise can either be unstructured (identity covariance) or structured following a multivariate Gaussian model.
We take a fully-trained base model and extracted the layer activations given unperturbed inputs $x_L$ and modified inputs $x'_L$.
From these, we computed the $n_L \times n_L$ empirical covariance matrix $\Cfull$ of the differences $x'_L - x_L$ over all minibatches of the training set and added $10^{-4}$ to the diagonal for stability.

We then add Gaussian noise with covariance $C$ to the activations at layer $L$ using the Cholesky decomposition.
We took either the 
full covariance $C = a \Cfull$, 
diagonal $C =\Cdiag = a \, \mathrm{diag}(\Cfull)$, 
or identity $C = aI$.
To ensure the strength of the noise was consistent across conditions, we multiplied the covariance matrix $C$ by a constant $a$ so that the {\em normalized trace}, or variance per dimension,
\begin{equation}
    \tr(C) = \frac{1}{n_L} \mathrm{Tr}(C)
\end{equation}
is fixed at a given scale.
``No $\tr$'' means we didn't perform trace normalization.
Then the weights of all layers up to and including layer $L$ were frozen, and the later layer weights were optimized to minimize loss on the training set for 10 epochs (Fig.~\ref{fig:supp_learning_curve} shows learning curves) using unperturbed data.

\subsection{Experimental Setup}

To compare the usefulness of the structured vs.\ unstructured noise, we evaluated the robustness of different noisy models to modified inputs.
Noisy models were trained using the procedure described above and then tested by delivering manipulated images from the test set that was unseen during training.
Unless otherwise specified, we match the type and strength of image manipulations (attacks, spatial transformations, etc.) used to create the covariance $C$ to those during testing.
We use the second convolutional layer as the default noisy layer, although we do show results for varying $L$ below.
To capture network and trial variability, we ran each experiment 10 times using models trained from independent initial weights and noise samples.
For experiments requiring the use of non-deterministically modified data (e.g.\ adversarial attacks or random perturbations), a fresh dataset was generated for each trial.
Plot shading indicates bootstrap 95\% confidence intervals.

\section{Results}

Injecting structured noise into the activations of a neural network improves the model's robustness against adversarial attacks, partial obstructions, and naturalistic image modifications, as shown in Fig.~\ref{fig:ItWorks}.
For adversarial attacks, while any noise offers some benefit, full covariance offers significantly better performance than identity or diagonal once $\varepsilon$ is large. 
For AutoPGD with $\varepsilon=0.16$6 and trace scale $\tr = 2.0$, the mean performance of models with full covariance noise was 0.55, compared to 0.24 for models with identity covariance noise, 0.34 for $\tr = 2.0$ diagonal covariance noise, and 0.00 for no noise. 
We found similar results for other white-box adversarial attacks (see Tab.~\ref{tab:results}).  
In contrast, the benefit of full covariance noise is significantly less pronounced for Square attack, and only appears for highest $\varepsilon$.

Full covariance offers most pronounced benefits for motion blur (Fig.~\ref{fig:ItWorks}).
At strength 4 and $\tr = 0.5$, models with full covariance noise had a test accuracy of 0.64, compared to 0.29 for both diagonal and identity noise, and 0.30 for no noise. 
In contrast with AutoPGD, identity and diagonal covariance perform the same as baseline noiseless network.
There is a similar pattern in the effects of structured noise on robustness against a wide range of naturalistic image modifications (Tab.~\ref{tab:results}).
Full covariance generally offers the best performance with identity and diagonal noise offering limited robustness to noisy modifications (Gaussian and impulse noise, snow), and little to no benefit for the other modifications/transformations.
One notable exception is the elastic transform, where none of the methods succeed in defending. 

\subsection{Optimal Noise Strength for Robustness}

\begin{figure*}[ht]
\centering
\includegraphics[width=1.0\textwidth]{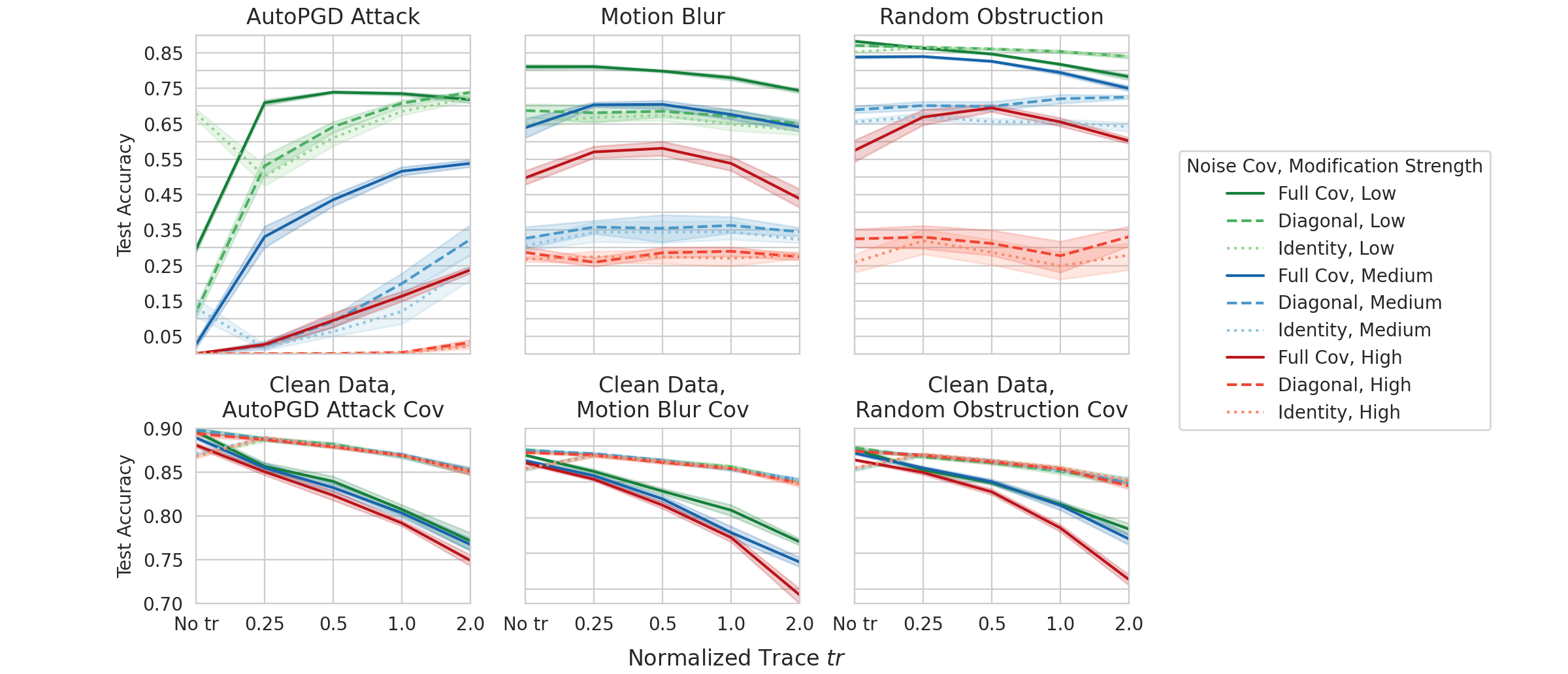}
\caption{
\textbf{Optimal noise strength depends on the image modification.}
Noise strength is set by the normalized trace $\tr$.
(Top Left) AutoPGD Attack: $\tr = 2.0$ outperforms all smaller trace values for medium and high strength attacks.
For no $\tr$, identity noise performs best, likely because it has the largest trace before normalization. 
Low, medium, and high modification strengths correspond to $\varepsilon$ values of $0.04$, $0.1$, and $0.18$ respectively. 
(Top Center) Motion Blur: $\tr = 0.25$, $\tr = 0.5$, and $\tr = 1.0$, provide similar performances, with no $\tr$ and $\tr=2.0$ underperforming.
Low, medium, and high modification strengths correspond to severity values of $1$, $3$, and $5$ respectively.
(Top Right) Random Obstruction:
No $\tr$ and $\tr=0.25$ perform slightly better for small and medium modification strengths, with $\tr=0.25$ and $\tr = 0.5$ performing best for high strength modifications.
Low, medium, and high modification strengths correspond to scale values of $0.4$, $1.0$, and $1.8$ respectively. 
(Bottom Row) Larger values of $\tr$ yield worse performance on clean data for all image alterations. 
See Fig.\ \ref{fig:supp_trace} in the appendix for the effect of noise strength for all modification types.}
\label{fig:trace}
\end{figure*}

The optimal value for normalized trace $\tr$ varies depending on the type of image modification used.
As shown in Fig.~\ref{fig:trace},  larger values of $\tr$ yield better robustness against mid to high $\varepsilon$ white-box adversarial attacks (AutoPGD, PGD, FGM).
For AutoPGD with $\varepsilon=0.10$, using full covariance noise with $\tr=2.0$ yields a performance of 0.55 95\% CI [0.54, 0.56] compared to 0.45 [0.43, 0.46] when $\tr=0.5$. 
However, this trend does not hold for Square attack or naturalistic image modifications, where instead lower $\tr$ values can result in better performance. 
As an example, for motion blur with severity=3 and full covariance noise, $\tr=2.0$ results in a performance of 0.64 95\% CI [0.62, 0.66] in comparison to 0.70 [0.69, 0.71] when $\tr=0.5$. 
The effect of $\tr$ on robustness against all types of image modification is shown in Fig.~\ref{fig:supp_trace} in the appendix.
Higher values of $\tr$ result in a significant decrease in performance on clean data (see Fig.~\ref{fig:trace}). 

\subsection{Structure is Most Effective in Early Layers}

\begin{figure*}[ht]
\centering
\includegraphics[width=1.0\textwidth]{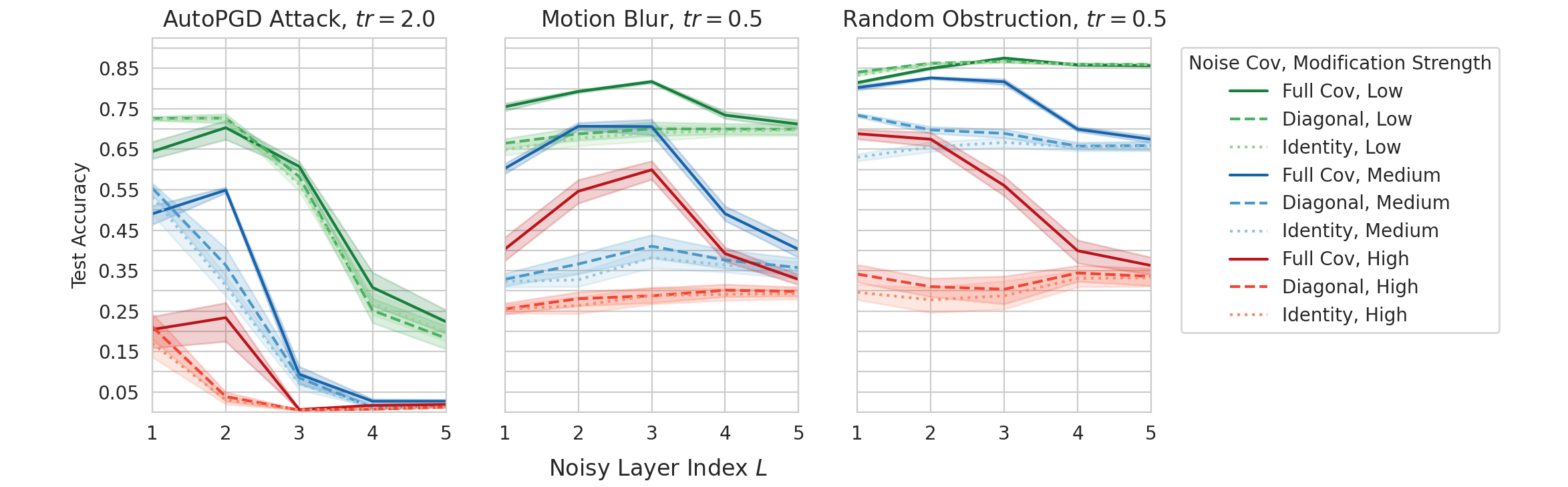}
\caption{\textbf{Noisy layer placement affects performance.} 
Mean test accuracy for all noisy layers $L$ and noise covariances. 
(Left) AutoPGD Attack. 
$L=1$ and $L=2$ offer the best performances.
For $L=2$, full covariance noise yields the highest accuracy for medium and high strength attacks.
When $L=1$, the difference in performances between covariances is smaller, with full covariance underperforming compared to diagonal for medium and low strength attacks. 
Low, medium, and high modification strengths correspond to $\varepsilon$ values of $0.04$, $0.1$, and $0.18$ respectively. 
(Center) Motion blur. 
$L=2$ and $L=3$ provide highest accuracies, with full covariance noise significantly outperforming diagonal and identity noise for all $L$.
Low, medium, and high modification strengths correspond to severity values of $1$, $3$, and $5$ respectively. 
(Right) Random obstruction. 
For high strength modifications, $L=1$ and $L=2$ offer the best performance, whereas for medium strength modifications $L=2$ and $L=3$ are preferable. 
Low, medium, and high modification strengths correspond to scale values of $0.4$, $1.0$, and $1.8$ respectively. 
See Fig.\ \ref{fig:supp_layers} in the appendix for the effect of layer placement for all modification types.}
\label{fig:layers}
\end{figure*}

The position of the noisy layer has a significant impact on performance, with the optimal noisy layer varying depending on the type and strength of image alteration.
As seen in Fig.~\ref{fig:layers}, the optimal noisy layer is always one of the convolutional layers (1--3), with the fully connected layers (4--5) under-performing.
The last fully connected layer was excluded from this experiment because adding noise to its activations is equivalent to adding noise directly to the model's output. 

When using the AutoPGD attack with $\tr=2.0$, layers $L=1$ and $L=2$ yield the best results.
Full covariance noise outperforms identity and diagonal noise for all $\varepsilon > 0.04$ with $L=2$.
When $L=2$ and $\varepsilon=0.10$, full covariance noise provides a performance of 0.54 95\% CI [0.53, 0.55], compared to 0.37 [0.35, 0.38] for diagonal and 0.26 [0.23, 0.29] for identity.
However, things change when $L=1$: 
Full covariance is best only at the highest attack strength ($\varepsilon = 0.2$), with diagonal noise generally yielding the best accuracy when $\varepsilon < 0.2$. 
For $L=1$ and $\varepsilon=0.10$, full covariance noise yields an accuracy of 0.49 95\% CI [0.47, 0.50], diagonal noise 0.56 [0.55,0.57], and identity noise 0.50 [0.44,55]. 

For motion blur with $\tr = 0.5$, $L=3$ provides the best performance with $L=2$ yielding the same or slightly lower accuracies.
As shown in Fig.~\ref{fig:layers}, full covariance noise consistently provides significantly better performance than identity or diagonal noise, regardless of the noisy layer selected. 
The effect of $L$ on robustness against all types of image modification is shown in Fig.~\ref{fig:supp_layers} in the appendix.

\subsubsection{Gaussian Augmentation}

We compared our method to to Gaussian data augmentation \citep{zantedeschi_efficient_2017}, which adds unstructured noise to the input (equivalent to identity noise at $L=0$). 
The standard deviation of the noise is set by $\sigma$. 
Note that we only evaluate the effects of Gaussian augmentation against adversarial attacks, excluding naturalistic image modifications.
We found that for low to moderate attack strengths, Gaussian noise with a standard deviation of $0.45$ provided comparable performance to structured noise in later layers, however full covariance noise at $L=2$ offered better performance against high strength attacks (see supplemental Fig.~\ref{fig:supp_gauss_aug}).
When augmentation with $\sigma=0.25$ is combined with full covariance noise with $\tr=2.0$ injected at $L=2$, performance improves compared to either method alone, particularly against high strength attacks. See supplemental Fig.~\ref{fig:supp_covadv_gauss_aug} for details. 

\subsection{Noise Covariance Transferability}

\begin{figure*}[t!]
\centering
\includegraphics[width=1\textwidth]{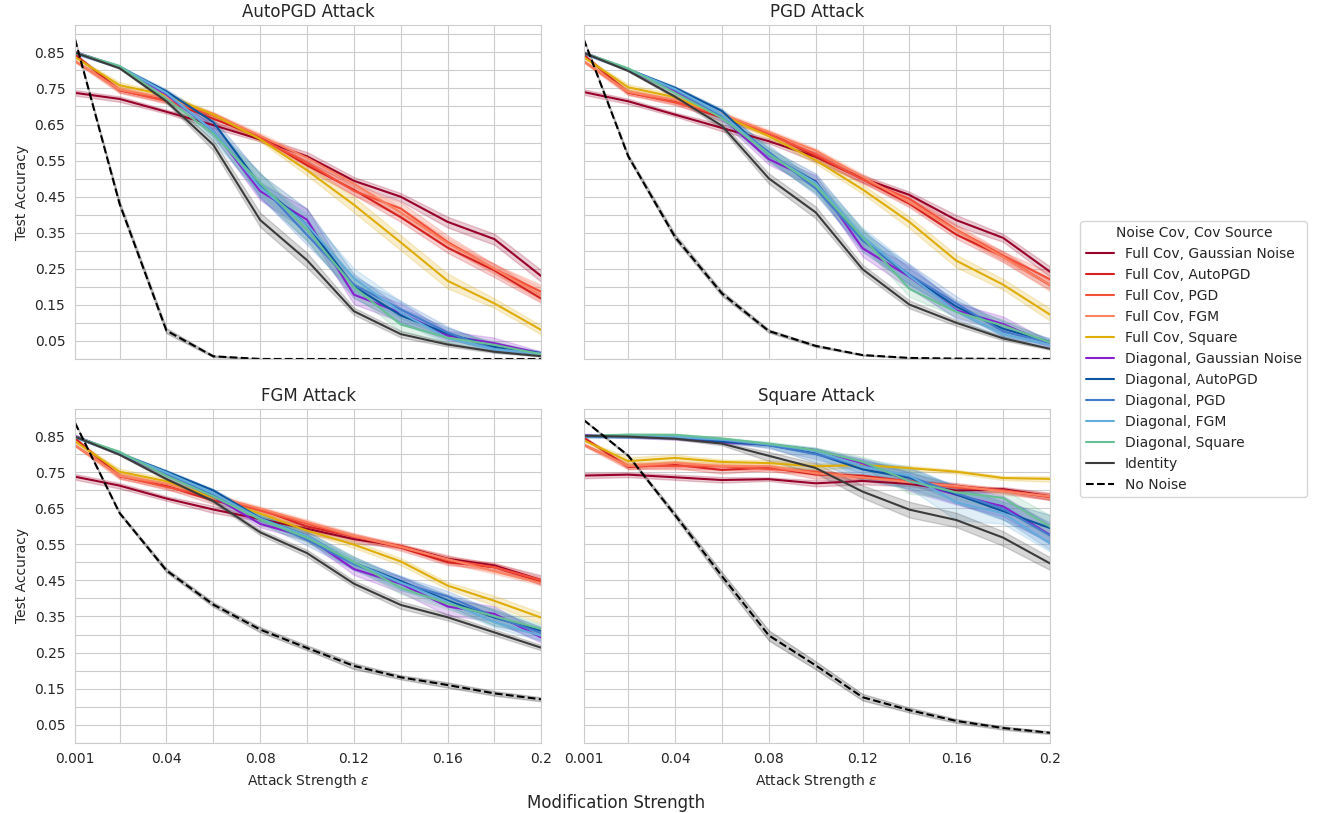}
\caption{\textbf{Noise covariances from many sources offer similar adversarial robustness.} 
Each subplot shows how models with a variety of noise covariances perform against a different attack over a range of  attack strengths (${\varepsilon}$).
Across all subplots, the difference in performance between white box attack (AutoPGD, PGD, and FGM) covariances is negligible.
(Top Left) AutoPGD attack.  Full covariance noise provides the best accuracy for all $\varepsilon > 0.04$.
Gaussian noise covariance provides slightly lower accuracies when $\varepsilon < 0.8$, but the highest accuracy by a narrow margin when $\varepsilon > 0.12$.
(Top Right) PGD Attack.  Similar to AutoPGD, with a smaller difference in performance between models with a Gaussian noise covariance and white box attack covariances.
(Bottom Left) FGM Attack. Gaussian noise covariance causes models to slightly underperform against low $\varepsilon$ attacks and does not provide a significant benefit for larger $\varepsilon$ attacks.
(Bottom Right) Square Attack. For high values of $\varepsilon$, full covariance noise derived from the Square attack yields the best result.
Diagonal noise from any source and identity noise offering better performance for lower values of $\varepsilon$.
Models with Gaussian noise covariance are less robust against this attack.}
\label{fig:covadv_equivalent}
\end{figure*}

The robustness gained from noise with a given covariance is not limited to only the attack used to generate the covariance.
Models with noise covariances derived from white box attacks (AutoPGD, PGD, and FGM) have remarkably similar performance against all tested adversarial attacks, as seen in Fig.~\ref{fig:covadv_equivalent}.
When evaluating robustness against white-box attacks, models with a covariance derived from a black-box attack (Square attack) perform similarly for lower values of $\varepsilon$, but at higher values their performance is notably lower.
Injecting noise with a covariance derived from the base model's activations in response to images augmented with Gaussian noise provides comparable adversarial robustness, particularly against white-box attacks.
For white-box attacks, models with Gaussian noise source covariance slightly underperform compared to adversarial source covariances for lower values of $\varepsilon$.
For AutoPGD with $\varepsilon = 0.04$ and full covariance noise, Gaussian noise covariance results in 0.69 95\% CI [0.68, 0.69] accuracy compared to 0.72 [0.71, 0.72] for AutoPGD covariance. 
The inverse is true for higher values of $\varepsilon$ against AutoPGD and PGD attacks: Although the effect is small, Gaussian noise covariance models are the most robust.
For AutoPGD, at $\varepsilon = 0.18$ and full covariance noise, Gaussian noise covariance yields an accuracy of 0.33 95\% CI [0.32, 0.35] compared to 0.25 [0.23, 0.26] for AutoPGD covariance. 

Noise covariances derived from other naturalistic image modifications also provide some level of adversarial robustness, but Gaussian noise is the only one that results in comparable performance to adversarially derived covariances (not shown). 
Impulse noise and snow provide lesser benefit, while others provide little to none.

As shown in Fig\. ~\ref{fig:covadv_equivalent}, model performance is largely determined by whether full covariance or diagonal covariance was used.
Full covariance provides the best robustness at higher values of $\varepsilon$, but diagonal noise offers the best robustness to low $\varepsilon$ attacks.
Identity and diagonal noise covariances result in very similar performance against low $\varepsilon$ attacks, but as $\varepsilon$ increases, identity noise becomes decreasingly effective. 

We evaluated whether robustness gains from structured noise persist when the noise covariance is derived from a different naturalistic image modification than the one used at evaluation time (Fig.~\ref{fig:mismatch_trans}).
In contrast to adversarial attacks, robustness to these modifications is largely non-transferable.
For most naturalistic modifications, mismatched noise covariances provide little to no benefit over models with no injected noise.
Several corruptions exhibit strong task specificity including random obstruction, rotation, motion blur, and perspective transformations.
A limited degree of transferability exists among corruptions that add noise or modify the brightness of the image, with Gaussian and impulse noise the most compatible.
Similarly, brightness-derived covariance provided the best robustness to changes in brightness, with snow-derived covariance also offering substantial robustness and all other covariance sources providing little to no benefit. 

\begin{table*}[ht]
    \centering
    { \small
\begin{tabular}{|ll|lllll|}
\hline
\multicolumn{2}{|c|}{\textbf{Data Modification}}    & \multicolumn{5}{c|}{\textbf{Test Accuracy by Noise Model}}                                      \\ \hline
    \multicolumn{1}{|l|}{\textbf{Type}}                 & 
    \textbf{Strength} & 
    \multicolumn{1}{l|}{$\mathbf{\tr}$} & 
    \multicolumn{1}{l|}{\textbf{Full}}                 & 
    \multicolumn{1}{l|}{\textbf{Diagonal}}             & 
    \multicolumn{1}{l|}{\textbf{Identity}}             &
    \textbf{No Noise}             \\ \hline
    \multicolumn{1}{|l|}{AutoPGD}          & 0.16     & \multicolumn{1}{l|}{2.0}     & \multicolumn{1}{l|}{\textbf{0.33} {[}0.32,0.35{]}} & \multicolumn{1}{l|}{0.08 {[}0.06,0.09{]}} & \multicolumn{1}{l|}{0.04 {[}0.03,0.05{]}} & 0.00 {[}0.00,0.00{]}    \\ \hline
    \multicolumn{1}{|l|}{FGM}              & 0.16     & \multicolumn{1}{l|}{2.0}     & \multicolumn{1}{l|}{\textbf{0.50} {[}0.49,0.51{]}}  & \multicolumn{1}{l|}{0.39 {[}0.37,0.41{]}} & \multicolumn{1}{l|}{0.35 {[}0.33,0.37{]}} & 0.16 {[}0.15,0.17{]} \\ \hline
    \multicolumn{1}{|l|}{PGD}              & 0.16     & \multicolumn{1}{l|}{2.0}     & \multicolumn{1}{l|}{\textbf{0.35} {[}0.35,0.36{]}} & \multicolumn{1}{l|}{0.13 {[}0.12,0.15{]}} & \multicolumn{1}{l|}{0.09 {[}0.07,0.11{]}} & 0.00 {[}0.00,0.00{]}    \\ \hline
    \multicolumn{1}{|l|}{Square}           & 0.16     & \multicolumn{1}{l|}{2.0}     & \multicolumn{1}{l|}{\textbf{0.75} {[}0.74,0.76{]}} & \multicolumn{1}{l|}{0.71 {[}0.68,0.73{]}} & \multicolumn{1}{l|}{0.59 {[}0.56,0.63{]}} & 0.06 {[}0.05,0.07{]} \\ \hline
    \multicolumn{1}{|l|}{Elastic}          & 1.5      & \multicolumn{1}{l|}{0.50}   & \multicolumn{1}{l|}{0.77 {[}0.77,0.78{]}} & \multicolumn{1}{l|}{\textbf{0.79} {[}0.79,0.80{]}}  & \multicolumn{1}{l|}{\textbf{0.79} {[}0.78,0.79{]}} & \textbf{0.79} {[}0.79,0.80{]}  \\ \hline
    \multicolumn{1}{|l|}{Perspective}      & 1.5      & \multicolumn{1}{l|}{0.50}   & \multicolumn{1}{l|}{\textbf{0.72} {[}0.69,0.75{]}} & \multicolumn{1}{l|}{0.61 {[}0.59,0.62{]}} & \multicolumn{1}{l|}{0.61 {[}0.60,0.62{]}}  & 0.61 {[}0.60,0.63{]}  \\ \hline
    \multicolumn{1}{|l|}{Obstruction} & 1.5      & \multicolumn{1}{l|}{0.50}   & \multicolumn{1}{l|}{\textbf{0.78} {[}0.77,0.78{]}} & \multicolumn{1}{l|}{0.48 {[}0.45,0.50{]}}  & \multicolumn{1}{l|}{0.44 {[}0.41,0.46{]}} & 0.46 {[}0.44,0.48{]} \\ \hline
    \multicolumn{1}{|l|}{Rotate}           & 1.5      & \multicolumn{1}{l|}{0.50}   & \multicolumn{1}{l|}{\textbf{0.63} {[}0.62,0.63{]}} & \multicolumn{1}{l|}{0.31 {[}0.30,0.31{]}}  & \multicolumn{1}{l|}{0.30 {[}0.29,0.31{]}}  & 0.28 {[}0.28,0.29{]} \\ \hline
    \multicolumn{1}{|l|}{Brightness}       & 4        & \multicolumn{1}{l|}{0.50}   & \multicolumn{1}{l|}{\textbf{0.84} {[}0.84,0.85{]}} & \multicolumn{1}{l|}{0.61 {[}0.57,0.64{]}} & \multicolumn{1}{l|}{0.44 {[}0.38,0.49{]}} & 0.57 {[}0.53,0.61{]} \\ \hline
    \multicolumn{1}{|l|}{Contrast}         & 4        & \multicolumn{1}{l|}{0.50}   & \multicolumn{1}{l|}{\textbf{0.49} {[}0.45,0.53{]}} & \multicolumn{1}{l|}{0.18 {[}0.15,0.22{]}} & \multicolumn{1}{l|}{0.16 {[}0.13,0.18{]}} & 0.23 {[}0.17,0.29{]} \\ \hline
    \multicolumn{1}{|l|}{Gaussian Noise}  & 4        & \multicolumn{1}{l|}{0.50}   & \multicolumn{1}{l|}{\textbf{0.82} {[}0.81,0.82{]}} & \multicolumn{1}{l|}{0.70 {[}0.65,0.75{]}}  & \multicolumn{1}{l|}{0.62 {[}0.57,0.67{]}} & 0.59 {[}0.52,0.65{]} \\ \hline
    \multicolumn{1}{|l|}{Impulse Noise}   & 4        & \multicolumn{1}{l|}{0.50}   & \multicolumn{1}{l|}{\textbf{0.82} {[}0.81,0.83{]}} & \multicolumn{1}{l|}{0.72 {[}0.70,0.75{]}}  & \multicolumn{1}{l|}{0.69 {[}0.65,0.72{]}} & 0.57 {[}0.53,0.61{]} \\ \hline
    \multicolumn{1}{|l|}{Motion Blur}     & 4        & \multicolumn{1}{l|}{0.50}   & \multicolumn{1}{l|}{\textbf{0.64} {[}0.63,0.66{]}} & \multicolumn{1}{l|}{0.29 {[}0.28,0.31{]}} & \multicolumn{1}{l|}{0.29 {[}0.28,0.31{]}} & 0.30 {[}0.28,0.32{]}  \\ \hline
    \multicolumn{1}{|l|}{Snow}             & 4        & \multicolumn{1}{l|}{0.50}   & \multicolumn{1}{l|}{\textbf{0.79} {[}0.79,0.80{]}}  & \multicolumn{1}{l|}{0.56 {[}0.53,0.60{]}}  & \multicolumn{1}{l|}{0.49 {[}0.45,0.54{]}} & 0.50 {[}0.48,0.53{]}  \\ \hline
    \end{tabular}
    }
    \caption{
    \label{tab:results}
    {\bf Typical results for all modifications, moderate image modification strength.}
    Accuracy is presented as mean and 95\% confidence intervals for different noisy models with the highest mean accuracies in bold.
    Generally, the best accuracy results from using the full covariance.
    }
\end{table*}

\subsection{Comparison to Adversarial Training}

To compare the effectiveness of our model to standard adversarial training, we used AdversarialTrainer with PGD from the ART library. 
We found that standard adversarial training offers more adversarial robustness than injecting structured noise, with lower ratios of adversarial examples in the training data resulting in reduced robustness (Fig.~\ref{fig:supp_adv_train}). 
Against PGD with $\varepsilon=0.1$, adversarially trained models with ratio 0.8 had a mean accuracy of $0.74 \pm 0.01$ compared to $0.57 \pm 0.01$ for a full covariance model with $L=2$ and $tr=2.0$. 

\section{Discussion}

Across a range of experiments on a convolutional network (and ViT; see appendix), structured noisy layers yield greater robustness than unstructured noise.
This demonstrates that the geometry of the noise relative to the network's representations of the data has a significant impact on robustness.

Full covariance noise generally confers greater robustness than diagonal and identity noise for moderate to high strength image modifications, with the exception of the elastic transformation. 
Another exception is when noise is injected into the earliest convolutional layer; here diagonal and identity noise yield better adversarial robustness than full covariance noise for low to moderate strength adversarial attacks.
It's possible the attacks tend to target individual neurons rather than subspaces, and that we can defend against these vulnerable neurons with larger variance; however, we did not directly test this hypothesis.
By later layers, full covariance noise is best.
For non-adversarial modifications, full covariance noise still generally yields the best performance, compared to diagonal and identity, across all layer choices.
Many of these modifications (e.g.\ elastic, perspective) are nonlinear effects on the image space that could potentially be linearized in later layers and thus more easily defended against in those locations.
Overall, our results show the most significant benefits of noise in the early layers.

One could make the argument that noise being more beneficial in early layers contradicts observations from biological neural networks, where activity in peripheral neurons is significantly less variable than in cortical neurons.  
However, this assumes that early layers of a CNN can be used as an analogue for peripheral neural circuitry. 
Perhaps it is more reasonable to view the CNN in its entirety as a cortical neural circuit, where the periphery corresponds to the model input. 

Noise strength, set by $\tr (C)$, significantly impacts robustness. Smaller values of $\tr$ tend to be most effective, with the exception of white-box adversarial attacks where a larger $\tr$ yields better robustness. 
This suggests a trade-off between injecting sufficiently strong noise to counter white-box adversarial perturbations and preserving task-relevant signal under more naturalistic distortions.

When evaluating robustness to adversarial attacks, diagonal and identity noise offer better robustness at low attack strengths. 
This may be because the extra parameters contained in a full covariance matrix are unnecessary for robustness to low strength attacks. 
It is possible that full covariance noise overfits to low strength attacks, decreasing performance. 
Diagonal noise tends to yield significantly greater robustness than identity noise, with exceptions for some modifications at low strengths and for motion blur and rotation at all strengths.
A diagonal covariance enables the noise strength to vary by neuron, whereas identity noise results in the same strength across all neurons. 
When diagonal noise does not offer an advantage, it indicates that there is no benefit to allowing the noise to vary by neuron without correlations between neurons, which can only be provided by a full covariance matrix.

No single covariance source provides robust performance across all modifications, indicating that our technique for generating structured noise does not encode a universal invariance, but instead makes the model resilient to a limited set of related perturbations. 

Compared to Gaussian augmentation, structured noise provides more robustness to high strength adversarial attacks.
When we combine Gaussian augmentation with our method, we see better performance than for either method alone, at least with the tested parameters.
We did not evaluate Gaussian augmentation against naturalistic modifications, however we expect that it would have little to no benefit. 
It would be equivalent to adding identity noise to the inputs, and identity noise tends to be ineffective against non-adversarial modifications. 

There are important limitations to our work.
While adding structured noise is a biologically plausible strategy for improving robustness, our method relies on seeing both clean an modified inputs knowing their identities.
If structured noise plays a role in perception, its strength and shape is likely the result of an interaction between local learning rules and priors optimized over evolutionary timescales, not an explicit estimation from paired clean and corrupted inputs.
Local learning rules leading to robust noise structure is an interesting direction for future work.
Our method is outperformed by standard adversarial training (Fig.~\ref{fig:supp_adv_train}) when evaluating robustness against adversarial attacks. 
For low to medium values of $\varepsilon$, Gaussian augmentation also generally provides better protection against attacks. 
Our approach should be viewed as a proof of concept that uses a simple mechanism for demonstrating that appropriately structured noise can confer robustness, rather than an optimal strategy for discovering such covariances.
Further, we do not test the effects of multiple noisy layers or a large array of architectures or datasets (e.g.\ ResNets, although we did try ViT on CIFAR-10).
It would be interesting to consider adding noise to particular elements of the transformer module, for instance mimicking a ``stochastic'' attention mechanism.
We do not consider realistic neuronal noise such as would be found with stochastic spiking models.
The loss in performance on clean data when structured noise is injected suggests that we have not found the ideal shape, where the variance of the noise is minimized perpendicular to the decision boundary \citep{venkatesh_role_2024}. 

Future work could explore whether effective noise covariances can be obtained via a biologically plausible methods such as Hebbian-style rules combined with stochastic neurons \citep{eppler_representational_2026,scott_beyond_2021}.
One could also expose the model to a mixture of diverse image modifications, rather than a single type of modification, or consider multiple noisy layers and see if further benefits are possible.
Additionally, evaluating robustness against a broader range of black-box attacks would help clarify whether the observed behavior of Square attacks is an outlier or indicative of a distinct pattern between white-box vs.\ black-box attacks. 
It would also be interesting to investigate whether the most effective layers for noise injection correspond to those with the largest differences in activations or where the attacks directions can be most separated from signal by subspaces.

We have shown that structured stochasticity in internal representations can substantially improve robustness in artificial neural networks.
By injecting noise with a covariance derived from differences in activations between clean and modified inputs, we show that structured noise outperforms unstructured noise across a wide range of adversarial attacks and naturalistic image modifications, with its effectiveness depending on noise strength and the layer at which it is applied.
While robustness to naturalistic modifications is largely task-specific, the strong transferability of adversarially derived covariances suggests that these perturbations share common structure.
Our results support the view that stochasticity in activations encourages robust representations, drawing a computational parallel to the stochasticity observed in the brain.

\printbibliography[title={References}]

\appendix

\clearpage

\section{Appendix}

\subsection{Contributions}

RP wrote the code, performed experiments, and created the tables and figures.
KDH performed the ViT experiments and analysis.
PV provided code that was used to start the project.
RP, PV, SM, and KDH conceived of the project.
SM edited, and RP and KDH wrote the paper.
KDH supervised the project.

We are grateful to anonymous CCN 2026 reviewers for helpful suggestions, including the ViT architecture and idea of testing whether the layer with largest deviations under modification is the most effective noisy layer.

\subsection{LLM Use}

LLM-enabled software (Antigravity editor + Gemini 3 Flash) were used to assist with coding the ViT experiments and analysis.

\subsection{Supplemental Data and Experiments}

The code to run the experiments is available at our GitHub repo
\href{https://github.com/glomerulus-lab/structured-noise-robustness}{structured-noise-robustness}.

Additional figures and tables are provided here. 
These are also referenced in the main text, but for completeness we detail them in this appendix.

Table \ref{tab:mod_descriptions} lists the image modifications and their parameters that were used. 
We also show an example of the modified images for different attack strengths, corruption severity, and modification scale.

\begin{figure*}[t!]
\centering
\includegraphics[width=0.8\linewidth]{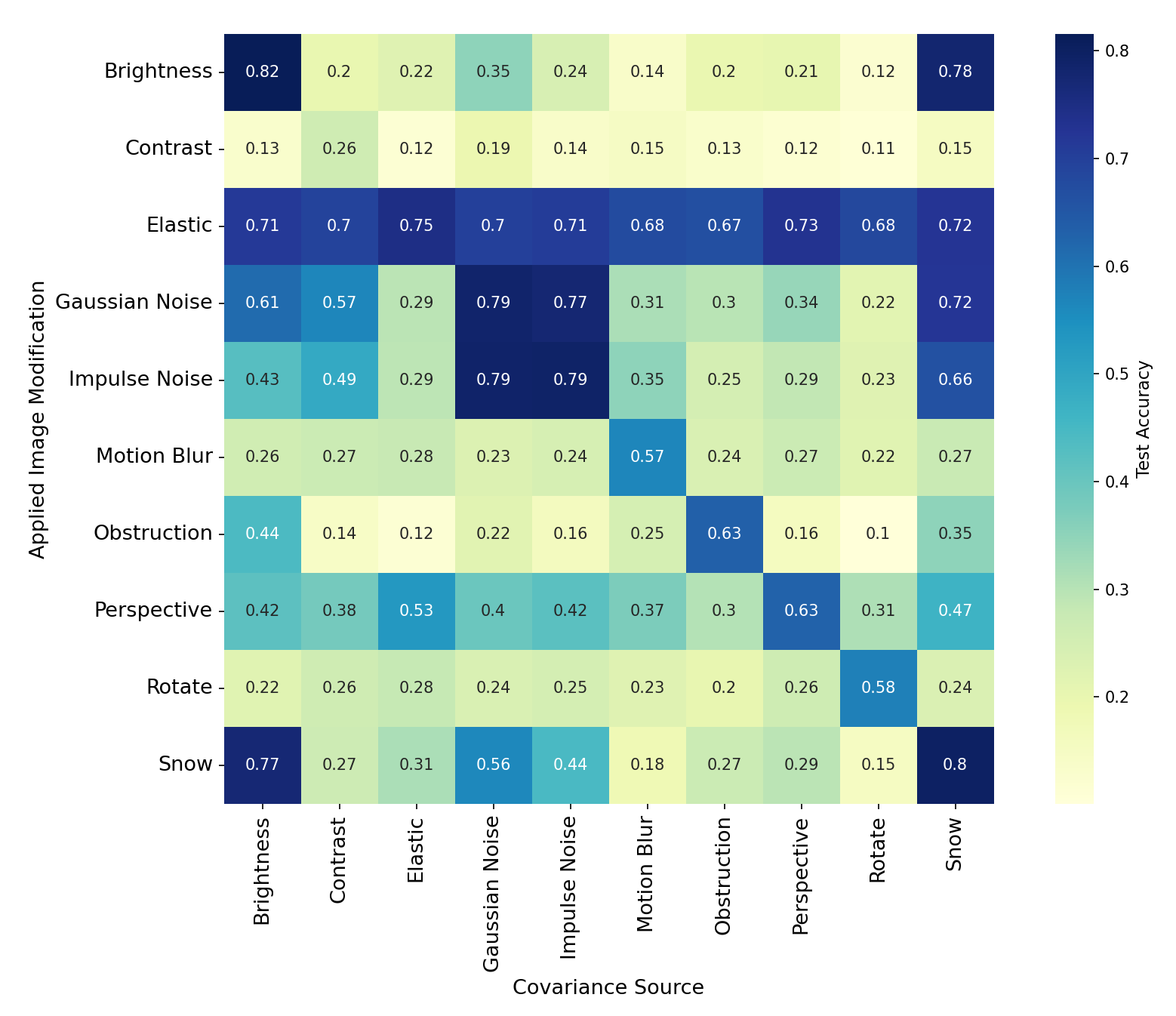}
\caption{
\textbf{Non-adversarial noise covariances have limited transferability.} 
Mean test accuracies for model with full covariance noise ($L=2$, $\tr=0.5$) over 10 runs; 95\% confidence intervals for all test accuracies are within $\pm0.05$. 
Maximum strength used for all image modifications (i.e.\ severity = 5 or scale = 2.0). 
Matched covariance source and modification (diagonal blocks) provides the best performance for all modifications, although elastic, Gaussian noise, and impulse noise have similar performance for other sources.
This is likely because structured noise confers little to no robustness to the elastic transformation and Gaussian noise and impulse noise display a high degree of cross-compatibility.
Brightness and snow-derived covariances also transfer well, but matched modification and covariance still provides the best average performance.}
\label{fig:mismatch_trans}
\end{figure*}

Fig.~\ref{fig:mismatch_trans} shows how accuracy varies when the covariance source does not match the applied modification, excluding adversarial modifications.
The heatmap shows that, aside from a few modifications such as Gaussian and impulse noise, most of the covariances do not transfer well.

\begin{figure*}[t!]
\centering
\includegraphics[width=0.9\linewidth]{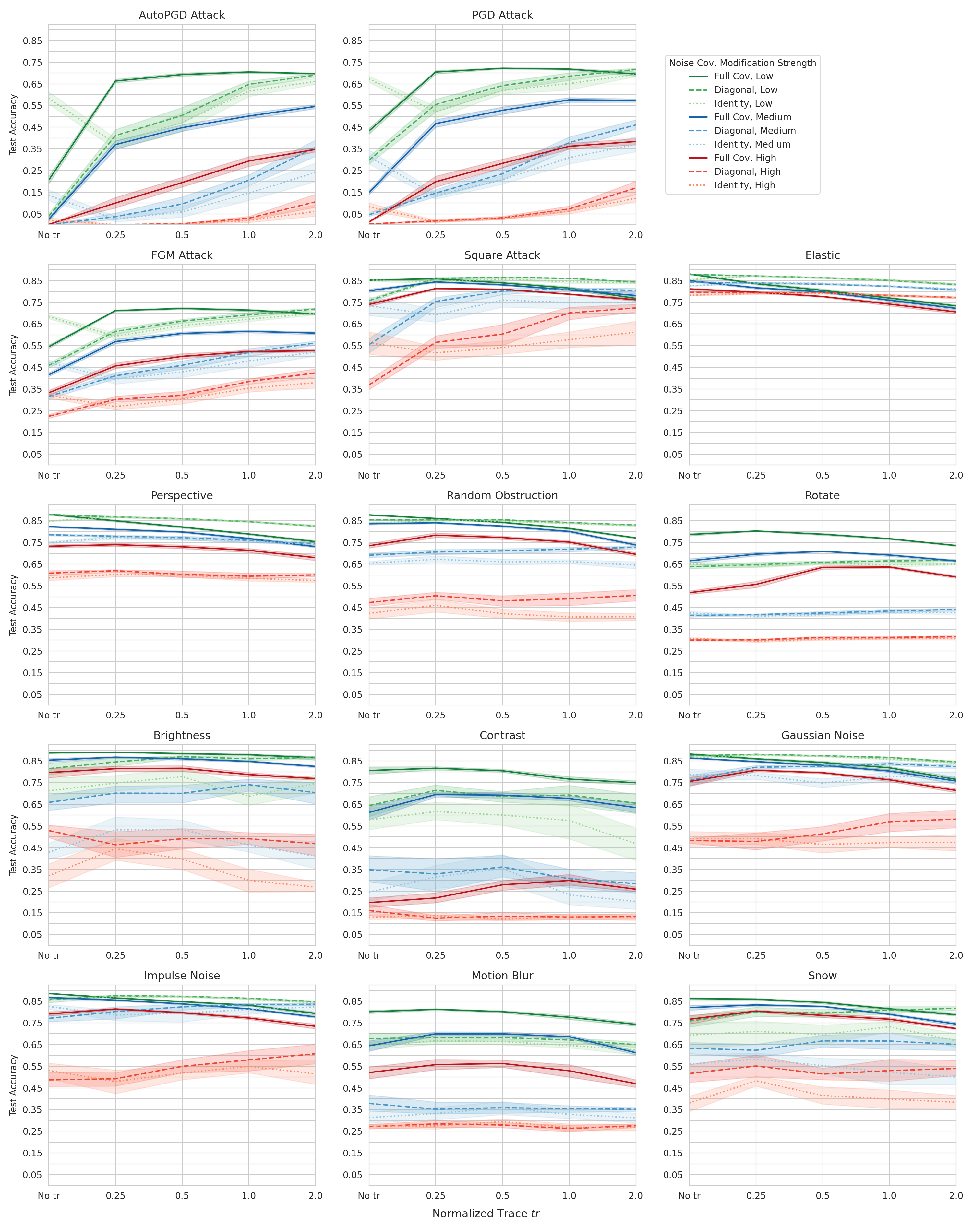}
\caption{
\textbf{Optimal noise strength depends on the image modification: all modifications.}
The trace scale $\tr$ controls the strength of the noise; no $\tr$) results in the lowest level.
More noise tends to provide better results for adversarial attacks, whereas for naturalistic modifications, lower levels of noise generally yield the greatest benefit. 
Low/medium/high strengths are $\varepsilon$ values of $0.05$/$0.10$/$0.15$ (adversarial), $0.5$/$1.0$/$1.5$ (obstruction, elastic, perspective, and rotate), or 1/3/4 (all other modifications).
}
\label{fig:supp_trace}
\end{figure*}

Fig.~\ref{fig:supp_trace} is a companion to Fig.~\ref{fig:trace}, showing the effect of noise strength $\tr$ on accuracy for all modifications and including 95\% confidence interval shading ($n=10$ for each data point).
We see that, across all modifications, larger $\tr$ leads to higher accuracy at large modification strength, while lower levels of noise are more effective with naturalistic modifications.

\begin{figure*}[t!]
\centering
\includegraphics[width=0.9\linewidth]{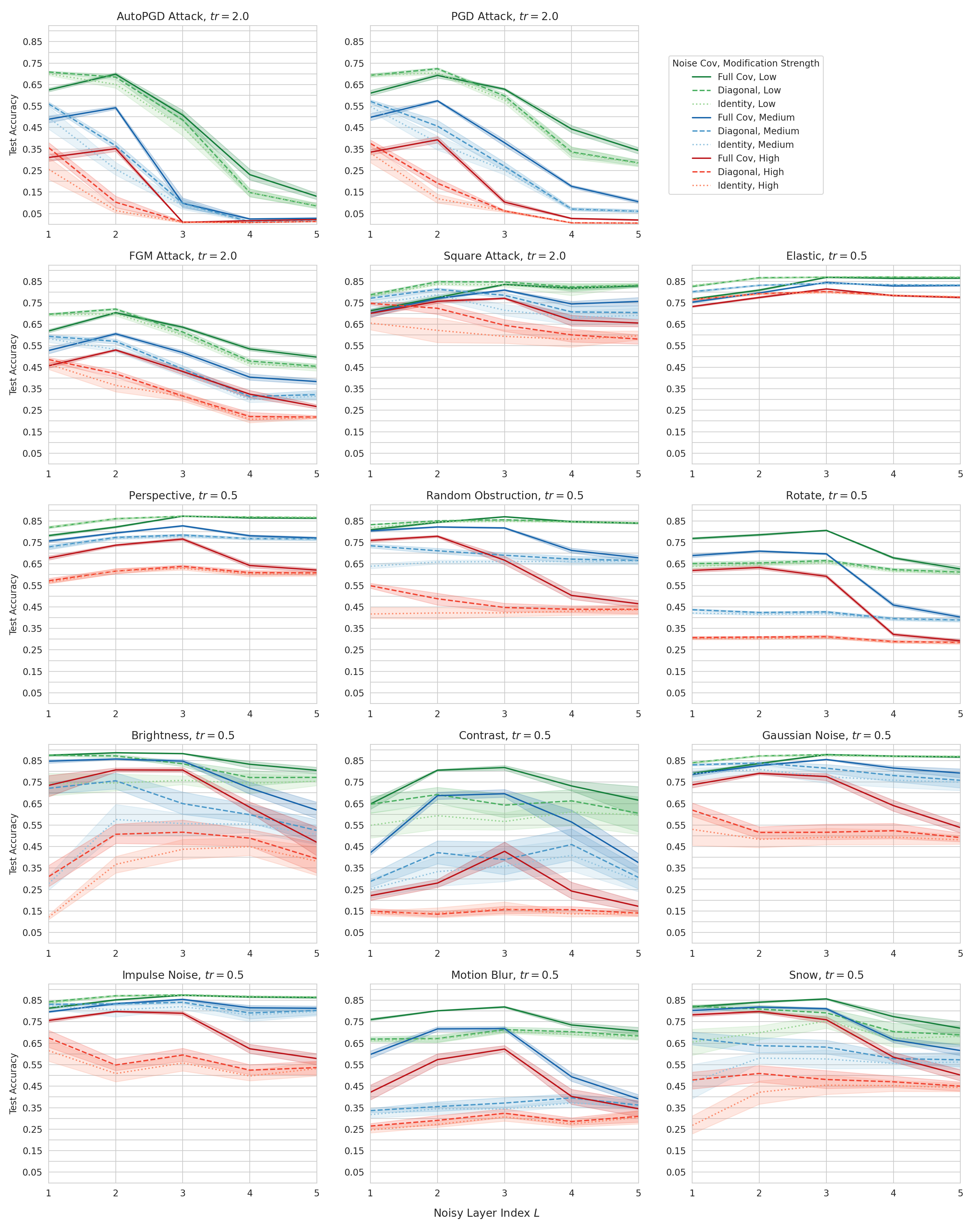}
\caption{
\textbf{Early noisy layer placement provides better robustness: all modifications.} 
The position of the noisy layer $L=1, 2, 3, 4, 5$ is shown by the line color.
Shading indicates 95\% confidence interval.
Injecting noise in one of the first 3 layers consistently provides better performance than injecting noise into later layers. 
For adversarial attacks, low, medium, and high modification strengths correspond to $\varepsilon$ values of $0.05$, $0.10$, and $0.15$ respectively. 
For random obstruction naturalistic image corruptions implemented with Torchvision transforms (e.g\. elastic, perspective, and rotate), low, medium, and high modification strengths correspond to scale values of $0.5$, $1.0$, and $1.5$ respectively. 
For all other image modifications, low, medium, and high modification strengths correspond to severity values of $1$, $3$, and $5$ respectively. }
\label{fig:supp_layers}
\end{figure*}

Fig.~\ref{fig:supp_layers} is a companion to Fig.~\ref{fig:layers} that shows how noisy layer position (shown by color) affects accuracy across modification strength for all modifications. 
For the adversarial attacks, earlier layers tend to lead to increased accuracy over later layers.
For other modifications, this effect is less pronounced.

\begin{figure*}[t!]
\centering
\includegraphics[width=\textwidth,trim={0 0 3cm 0},clip]{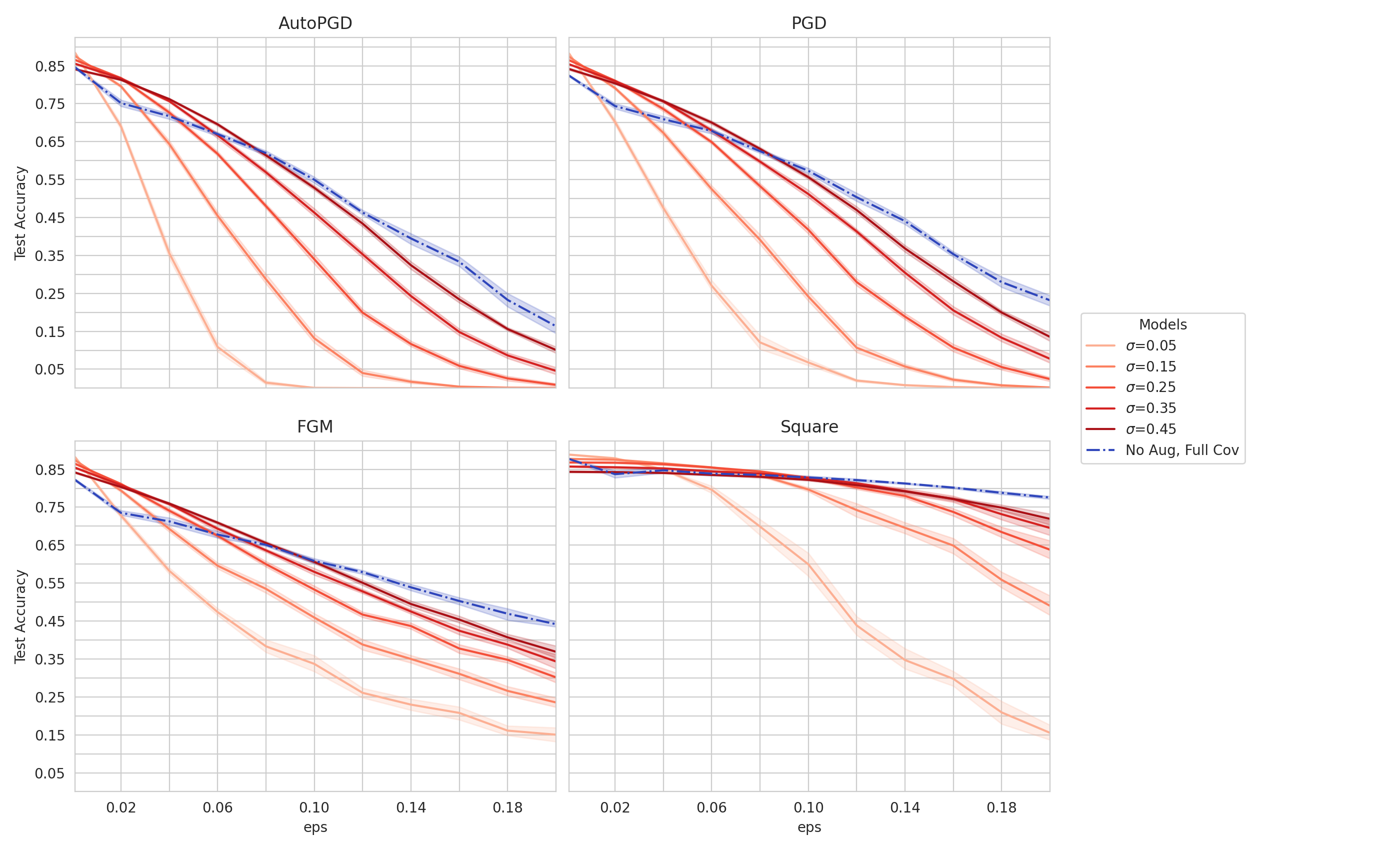}
\caption{
\textbf{Structured noise injection outperforms Gaussian augmentation at mid to high attack strengths}.
Here $\sigma$ sets the standard deviation of the Gaussian noise, with higher values providing greater robustness.
Shading indicates 95\% confidence interval.
A model trained with full covariance noise injection and no Gaussian augmentation is shown for comparison, with $\tr=0.5$ for Square attack and $\tr=2.0$ for all others.
}
\label{fig:supp_gauss_aug}
\end{figure*}

Fig.~\ref{fig:supp_gauss_aug} compares the effect of injecting noise to adding unstructured noise directly to the input image (known as Gaussian augmentation or randomized smoothing). 
For simplicity, just adversarial modifications are shown.
At larger modification strengths, structured noise added to the layers (our method, blue) leads to higher accuracy.

\begin{figure*}[t!]
\centering
\includegraphics[width=1\textwidth,trim={0 0 3cm 0},clip]{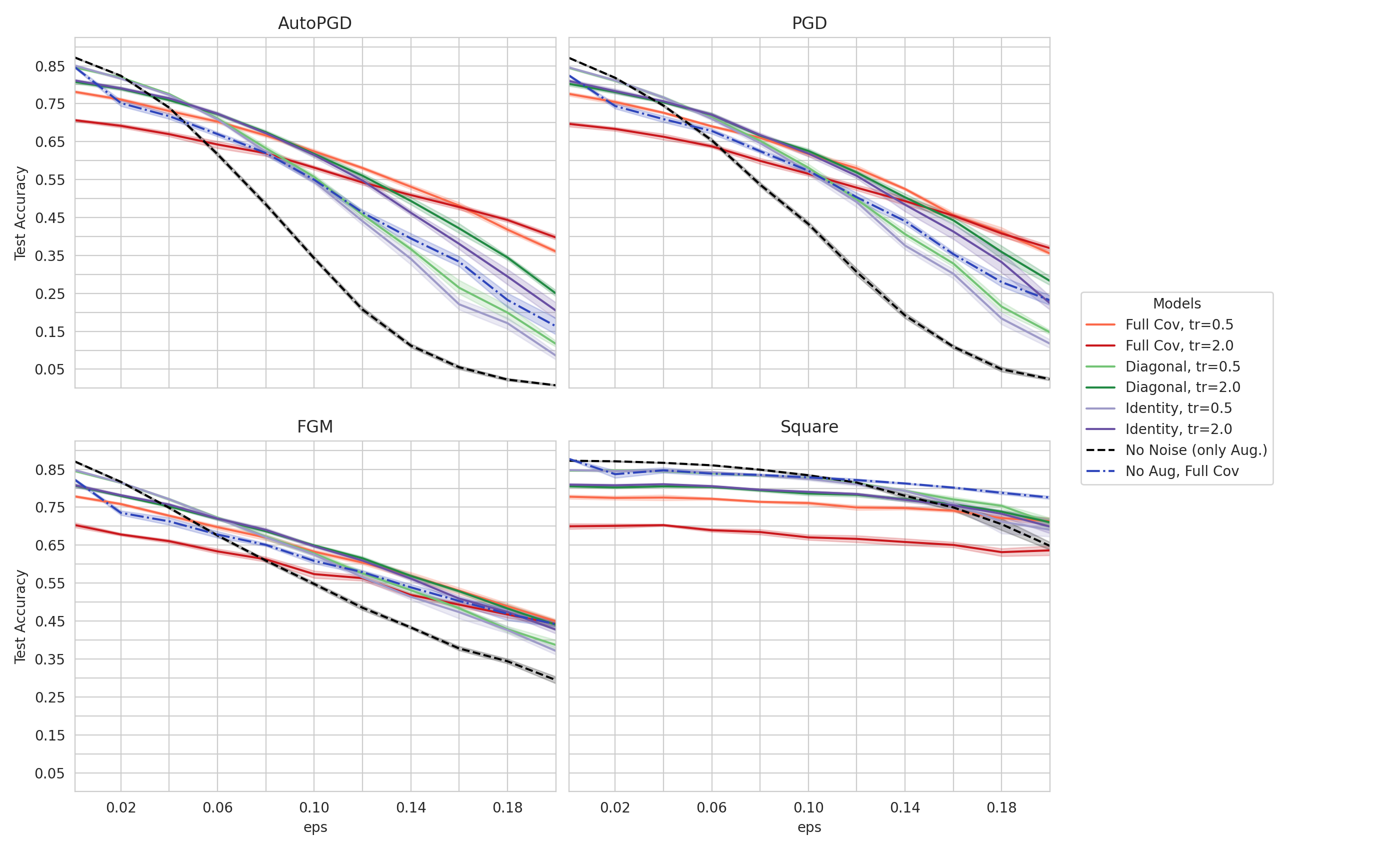}
\caption{
\textbf{Combining noise injection with Gaussian augmentation ($\sigma = 0.25$) provides better adversarial robustness than either method on their own, particularly for high $\varepsilon$ attacks.}
However, the difference in performance between structured and unstructured noise is notably smaller, with structured noise only providing a clear benefit for AutoPGD and PGD at high $\varepsilon$. 
Shading indicates 95\% confidence interval.
}
\label{fig:supp_covadv_gauss_aug}
\end{figure*}

Fig.~\ref{fig:supp_covadv_gauss_aug} shows how combining noise injection into later layers can be combined with Gaussian augmentation ($\sigma=0.25$) at the image layer to defend against adversarial attacks. 
While the ordering of the methods depends on attack and noise strengths, we see that combined noise in input and later layers (solid lines) may outperform noise in the input alone (dashed black) and in a later layer alone (dash-dotted blue).

\begin{figure*}[t!]
\centering
\includegraphics[width=1\textwidth]{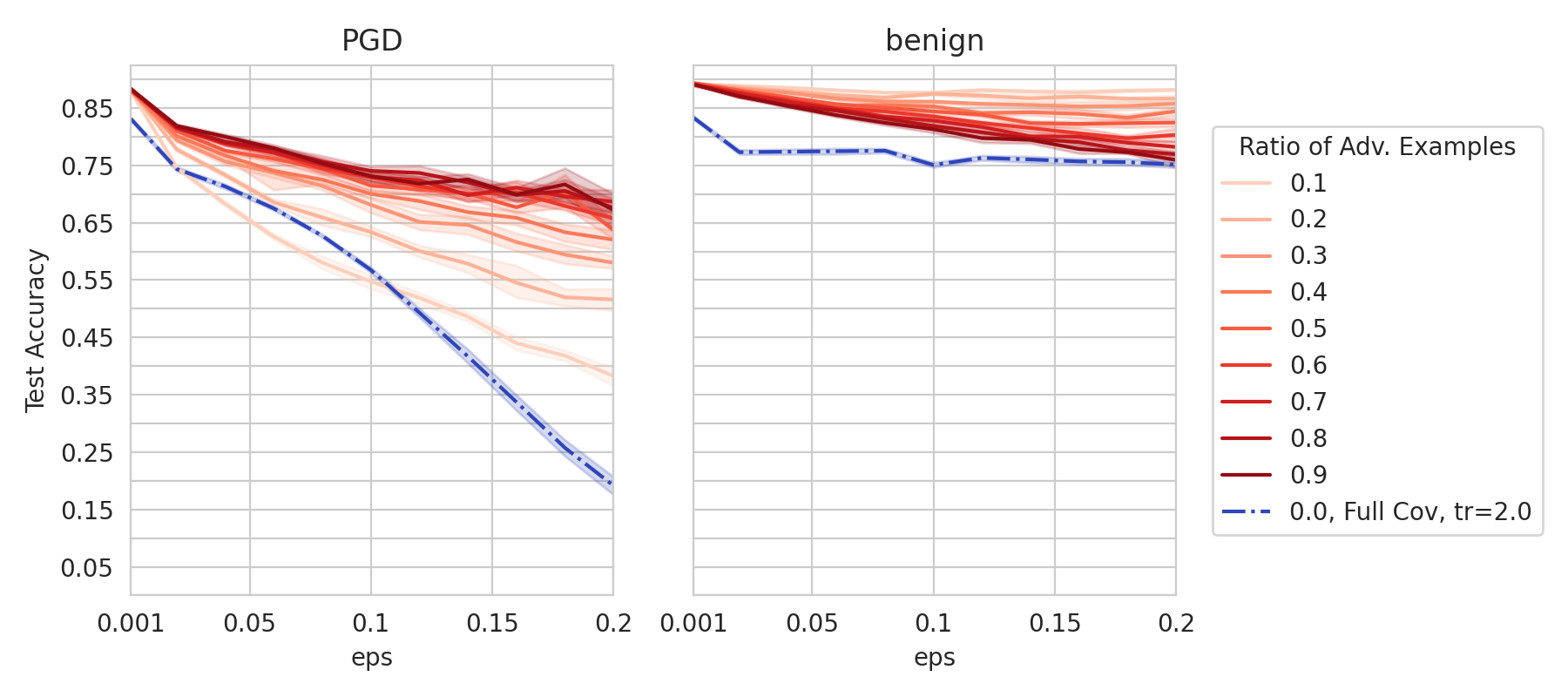}
\caption{
\textbf{Comparison to standard adversarial training with PGD using AdversarialTrainer from ART. }
Higher ratios of adversarial examples during training tend to provide better adversarial robustness, at the cost of worse performance on clean data. 
The performance of a model with no adversarial training with full covariance noise injection is shown for comparison in blue.
}
\label{fig:supp_adv_train}
\end{figure*}

Fig.~\ref{fig:supp_adv_train} compares adding noise to later layers with standard adversarial training, where the training examples and attacks were generated with PGD.
We see that adversarial training is generally the most effective way to defend against adversarial attacks. 
This is not unexpected, since it tunes all parameters of the model to defend against the adversarial data distribution.
On the other hand, shaping the noise in neural layers is simpler, could rely on less data, and could reflect mechanisms active in the brain that depend on only local information about layer activations.

\begin{figure*}[t!]
\centering
\includegraphics[width=0.9\linewidth]{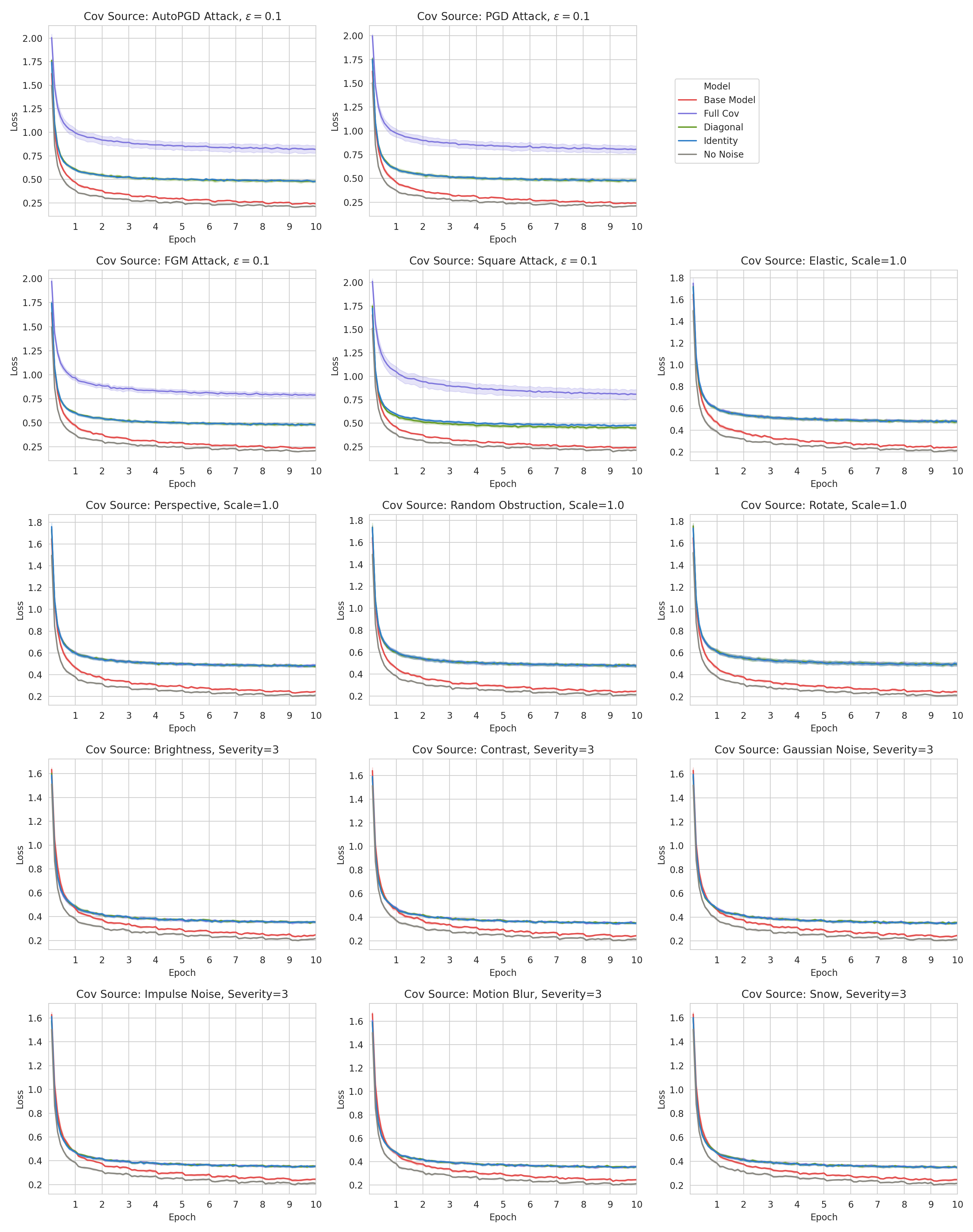}
\caption{
\textbf{Training curves indicate that 10 epochs of re-training is sufficient.} We train a base model, inject noise with a given covariance into the noisy layer, and retrain all following layers. For all non-adversarial modifications, the learning curves for full cov, diagonal, and identity noise are essentially identical. However  for adversarial attacks, models with full covariance noise have a consistently higher loss compared to diagonal and identity. The noise strength was set to $\tr=2.0$ for all adversarial attacks and $\tr=0.5$ for all other modifications. Shading indicates a 95\% confidence interval. 
}
\label{fig:supp_learning_curve}
\end{figure*}

Fig.~\ref{fig:supp_learning_curve} shows learning curves for the base models and noisy models across covariance sources.  
These curves indicate that 10 epochs of re-training is sufficient for the noisy models to converge. 
We found that the loss for full covariance, diagonal, and identity noise is remarkably similar when the covariance source is a naturalistic modification, however models with full covariance noise have significantly higher loss when the covariance is derived from an adversarial attack. 
Models with no noise (noise covariance matrix contains all zeros) that undergo the process of post "noisy layer" retraining have a slightly lower loss than the base model. 

\subsubsection{Vision Transformer Experiments}

To highlight the effect of neural variability in different model architectures and data, we applied the same methodology in a vision transformer \citep[ViT;][]{dosovitskiy_image_2021} classifying the CIFAR-10 tiny images dataset.
Our code was built from \href{https://github.com/Trusted-AI/adversarial-robustness-toolbox/blob/main/notebooks/huggingface_notebook.ipynb}{example code} provided by the ART package.
The pretrained {\tt WinKawaks/vit-tiny-patch16-224} model was loaded from Huggingface and finetuned on CIFAR-10 for 2 epochs, reaching accuracy $>$98\%.
Attacks were generated using the training split, degrading the undefended model accuracy to a few percent depending on attack strength.
We took the output of the first transformer block as the noisy layer (\texttt{vit.encoder.layer.0.output.dropout}).
Activation deviations were averaged over tokens/patch dimension to compute the covariance matrix, which had its trace rescaled as described in the main text.

\begin{figure*}[t!]
\centering
\includegraphics[width=.8\textwidth]{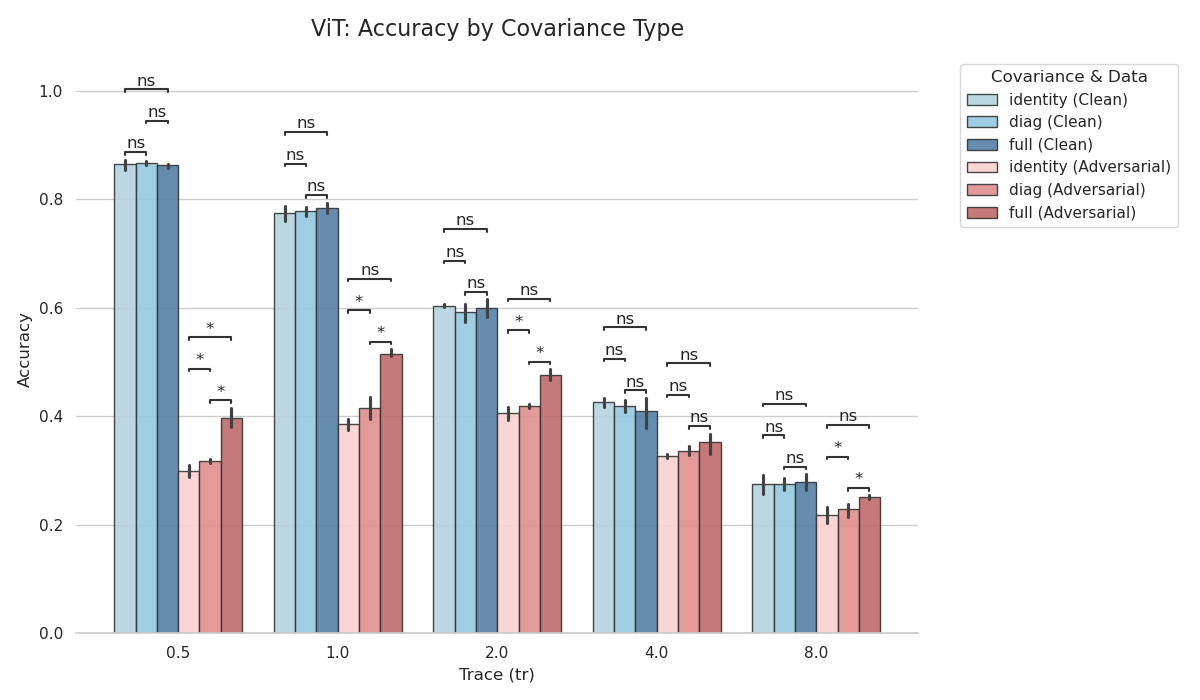}

\includegraphics[width=.8\textwidth]{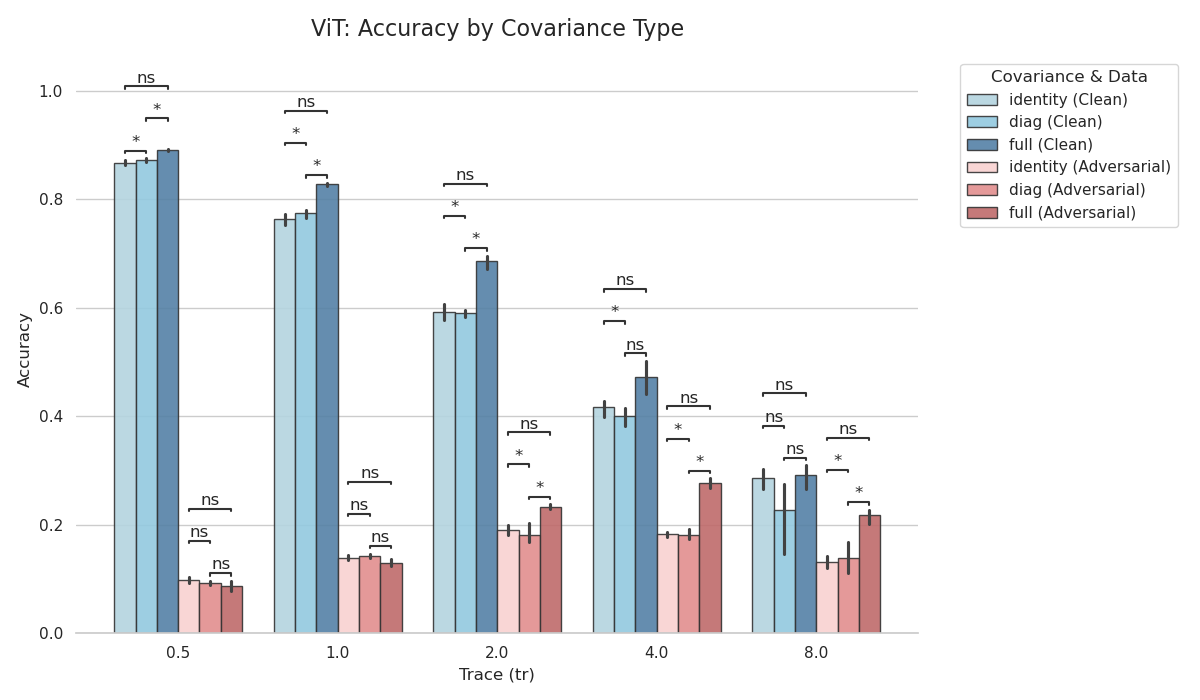}
\caption{
\textbf{Our method applied to vision transformer classifying CIFAR-10 data under PGD attack.}
Above: $\varepsilon = 4/255$, Below: $\varepsilon = 8/255$.
Similar to our convolutional network results, accuracy on clean data degrades with increasing $\tr$, and there is a sweet spot for adversarial robustness.
For the optimal values of $\tr$ at either attack strength ($\tr = 1$ for $\varepsilon = 4/255$; $\tr = 4$ for $\varepsilon = 8/255$), full covariance leads to significant improvements in adversarial accuracy.
Significance results follow Mann-Whitney U tests with * indicating $p < 0.05$; all data are $n=4$ independently fine-tuned runs (randomness arises from minibatch sampling and noise injection).
}
\label{fig:supp_vit}
\end{figure*}

Results are shown in Fig.~\ref{fig:supp_vit} for two attack strengths ($\varepsilon = 4/255$ and $8/255$) for $\tr = 0.5, 1, 2, 4, 8$ using identity, diagonal, and full covariance noise.
As with the convolutional network, structured noise may improve robustness on both clean and noisy data, and the optimal noise strength is larger for larger attack strength.
We did not investigate the effect of noisy layer location, other kinds of image modifications, transferability, or comparisons with adversarial training.

\begin{table*}[hb!]
{\small
\begin{tabular}{|l|c|c|c|}
\hline
\textbf{Image Modifications from Adversarial Robustness Toolbox}   
& \textbf{$\bm{\varepsilon}$=0.02} & \textbf{$\bm{\varepsilon}$=0.10} & \textbf{$\bm{\varepsilon}$=0.20} \\ \hline
AutoProjectedGradientDescent(eps=$\varepsilon$)
& \parbox[c]{1.3cm}{\includegraphics[width=1.3cm]{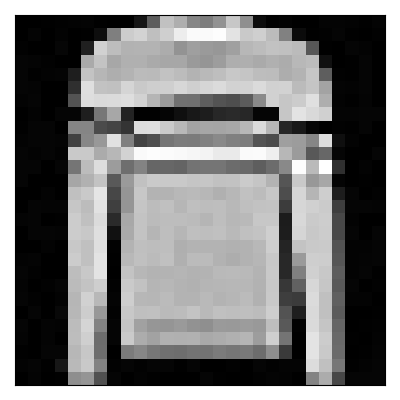}}
& \parbox[c]{1.3cm}{\includegraphics[width=1.3cm]{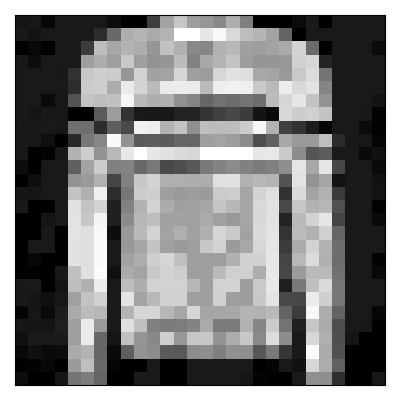}}
& \parbox[c]{1.3cm}{\includegraphics[width=1.3cm]{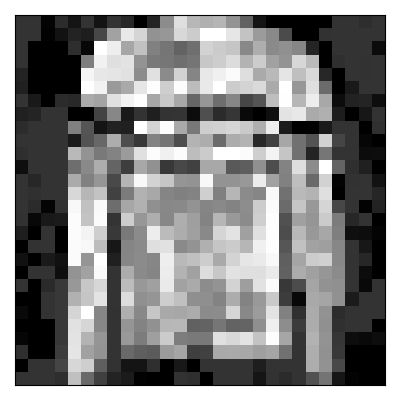}}           \\ \hline
FastGradientMethod(eps=$\varepsilon$)
& \parbox[c]{1.3cm}{\includegraphics[width=1.3cm]{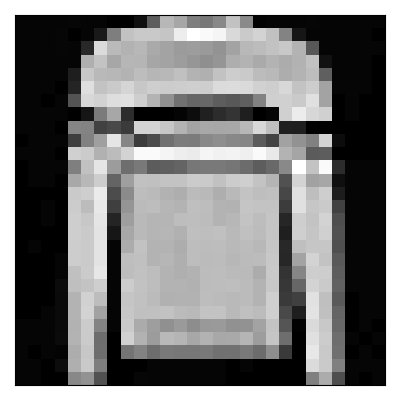}}
& \parbox[c]{1.3cm}{\includegraphics[width=1.3cm]{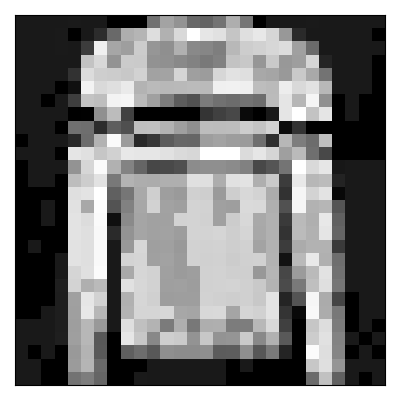}}
& \parbox[c]{1.3cm}{\includegraphics[width=1.3cm]{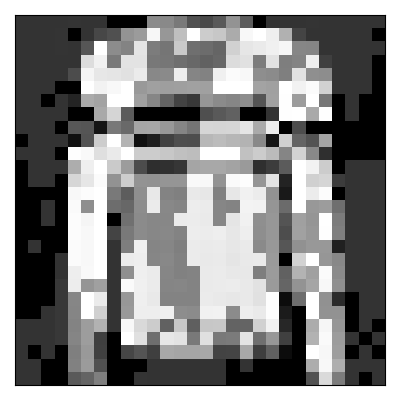}}           \\ \hline
ProjectedGradientDescent(eps=$\varepsilon$)
& \parbox[c]{1.3cm}{\includegraphics[width=1.3cm]{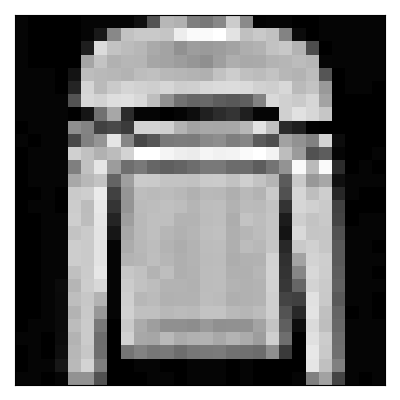}}
& \parbox[c]{1.3cm}{\includegraphics[width=1.3cm]{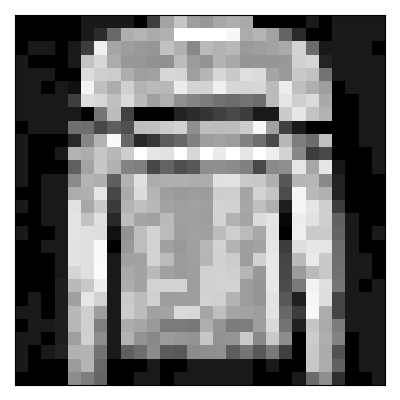}}
& \parbox[c]{1.3cm}{\includegraphics[width=1.3cm]{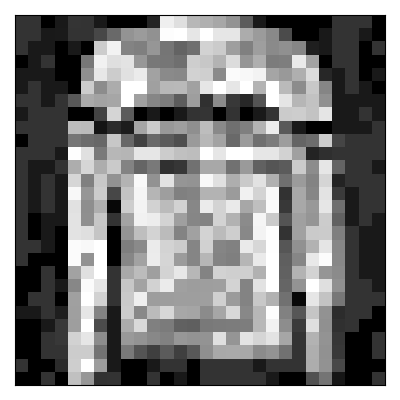}}           \\ \hline
SquareAttack(eps=$\varepsilon$)
& \parbox[c]{1.3cm}{\includegraphics[width=1.3cm]{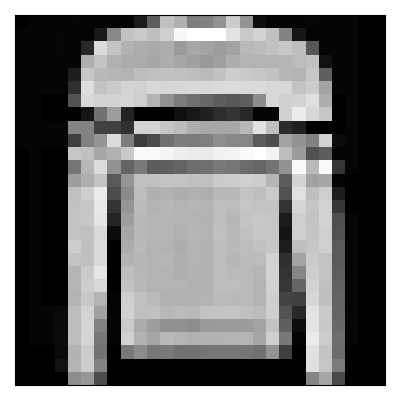}}
& \parbox[c]{1.3cm}{\includegraphics[width=1.3cm]{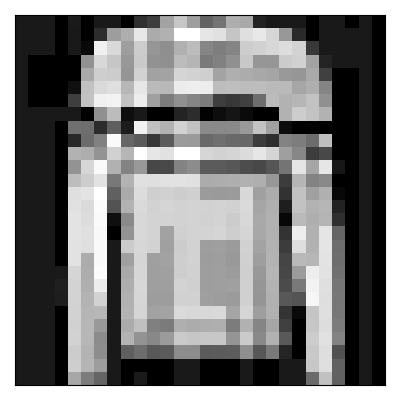}}
& \parbox[c]{1.3cm}{\includegraphics[width=1.3cm]{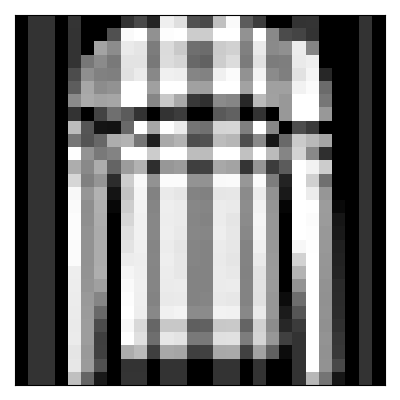}}           \\ \hline
\textbf{Image Modifications from imagecorruption}
& \textbf{severity=1} & \textbf{severity=3} & \textbf{severity=5} \\ \hline
corrupt(corruption\_name=``brightness'', severity=severity)
& \parbox[c]{1.3cm}{\includegraphics[width=1.3cm]{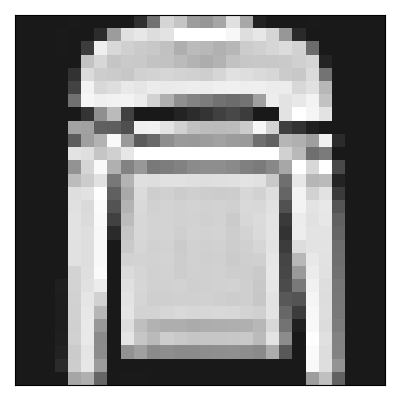}}
& \parbox[c]{1.3cm}{\includegraphics[width=1.3cm]{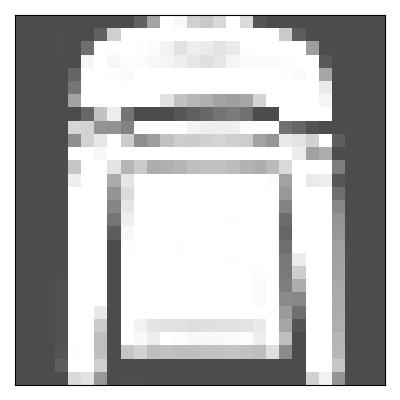}}
& \parbox[c]{1.3cm}{\includegraphics[width=1.3cm]{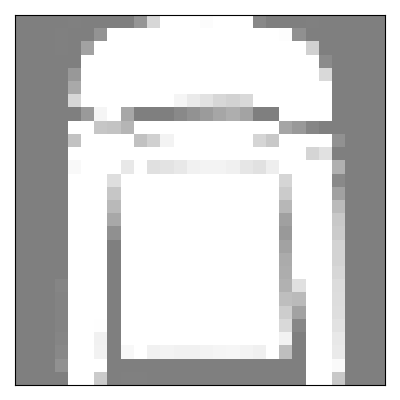}}           \\ \hline
corrupt(corruption\_name=``contrast'', severity=severity)
& \parbox[c]{1.3cm}{\includegraphics[width=1.3cm]{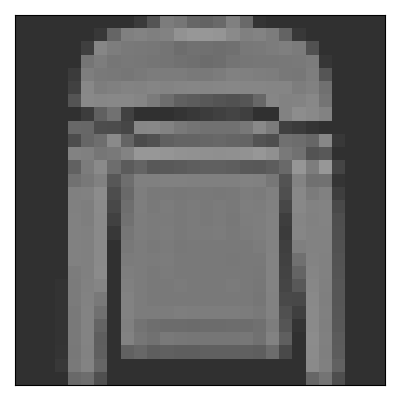}}
& \parbox[c]{1.3cm}{\includegraphics[width=1.3cm]{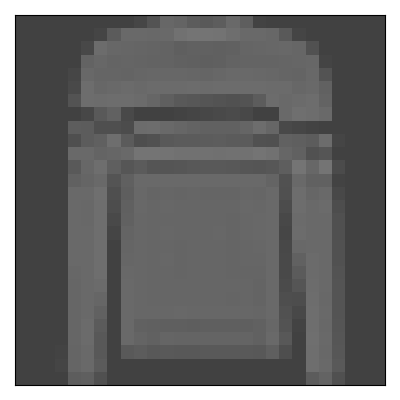}}
& \parbox[c]{1.3cm}{\includegraphics[width=1.3cm]{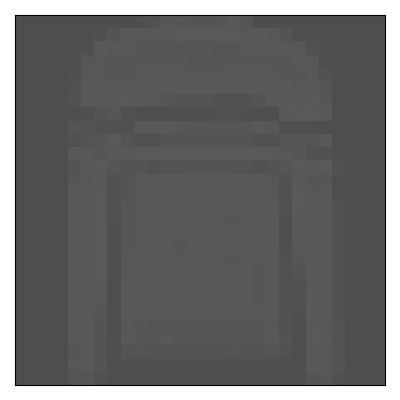}}           \\ \hline
corrupt(corruption\_name=``gaussian\_noise'', severity=severity)
& \parbox[c]{1.3cm}{\includegraphics[width=1.3cm]{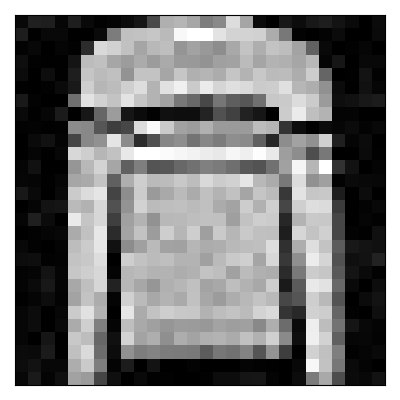}}
& \parbox[c]{1.3cm}{\includegraphics[width=1.3cm]{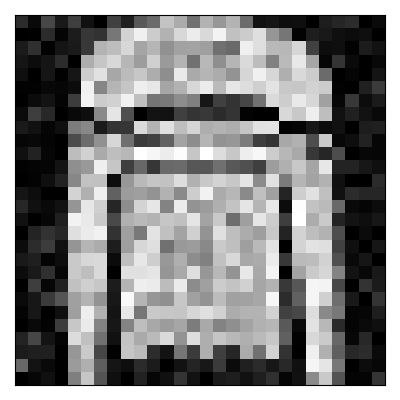}}
& \parbox[c]{1.3cm}{\includegraphics[width=1.3cm]{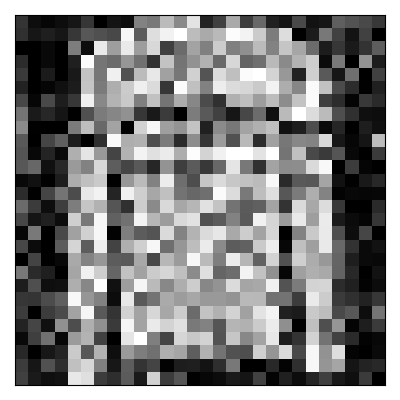}}           \\ \hline
corrupt(corruption\_name=``impulse\_noise'', severity=severity)
& \parbox[c]{1.3cm}{\includegraphics[width=1.3cm]{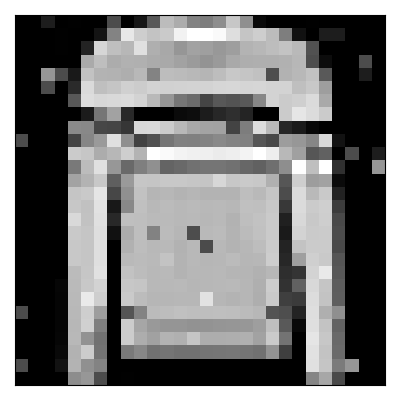}}
& \parbox[c]{1.3cm}{\includegraphics[width=1.3cm]{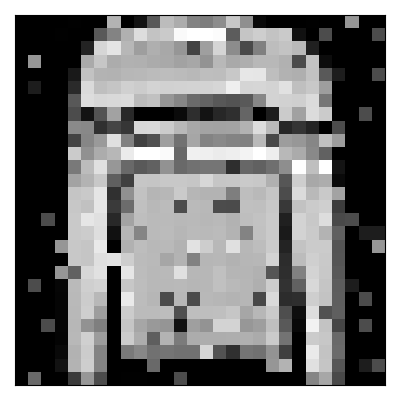}}
& \parbox[c]{1.3cm}{\includegraphics[width=1.3cm]{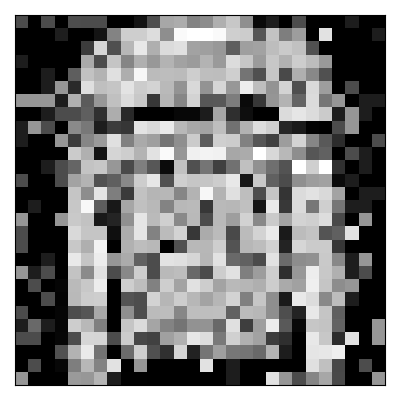}}           \\ \hline
corrupt(corruption\_name=``motion\_blur'', severity=severity)
& \parbox[c]{1.3cm}{\includegraphics[width=1.3cm]{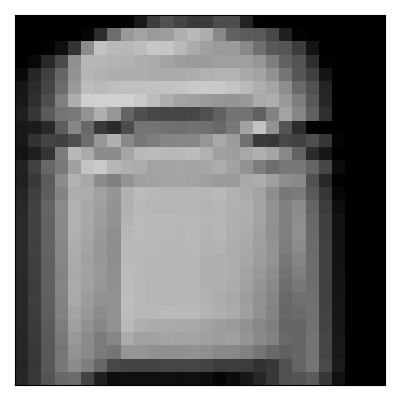}}
& \parbox[c]{1.3cm}{\includegraphics[width=1.3cm]{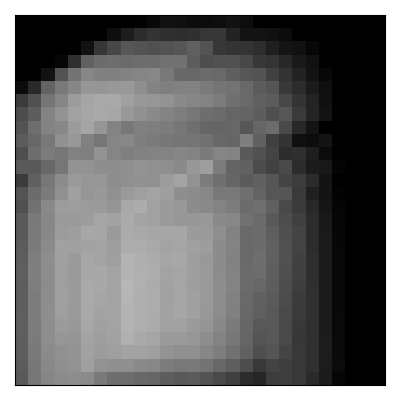}}
& \parbox[c]{1.3cm}{\includegraphics[width=1.3cm]{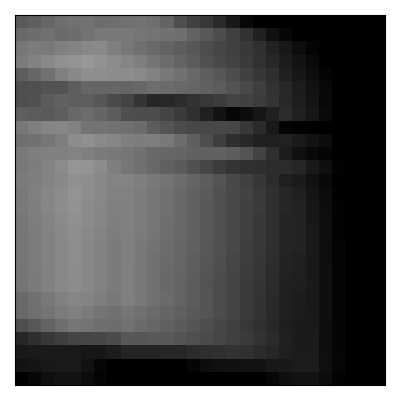}}           \\ \hline
corrupt(corruption\_name=``snow'', severity=severity)
& \parbox[c]{1.3cm}{\includegraphics[width=1.3cm]{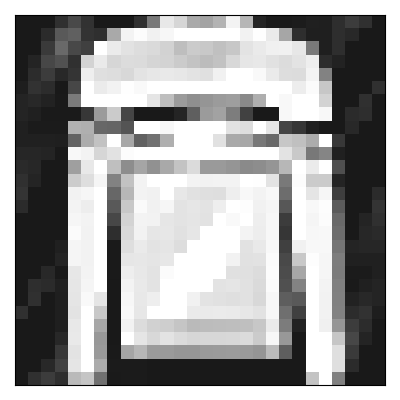}}
& \parbox[c]{1.3cm}{\includegraphics[width=1.3cm]{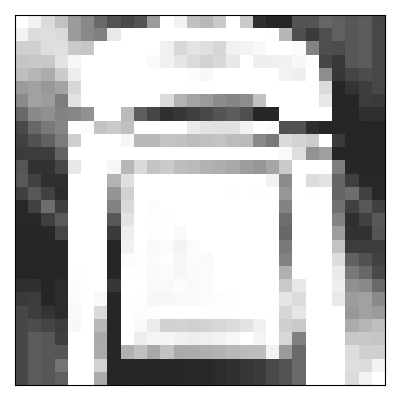}}
& \parbox[c]{1.3cm}{\includegraphics[width=1.3cm]{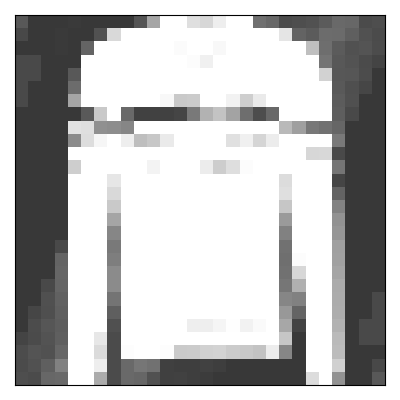}}           \\ \hline
\textbf{Image Modifications from torchvision.transforms.v2}
& \textbf{scale=0.2} & \textbf{scale=1.0} & \textbf{scale=2.0} \\ \hline
ElasticTransform(alpha = 35.0 * scale) 
& \parbox[c]{1.3cm}{\includegraphics[width=1.3cm]{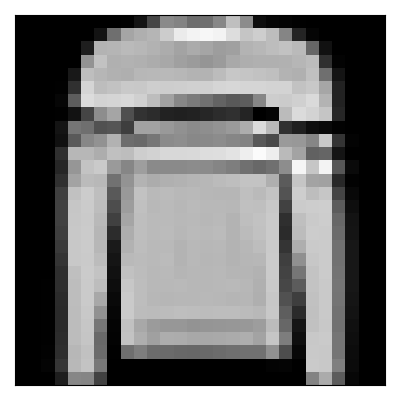}}
& \parbox[c]{1.3cm}{\includegraphics[width=1.3cm]{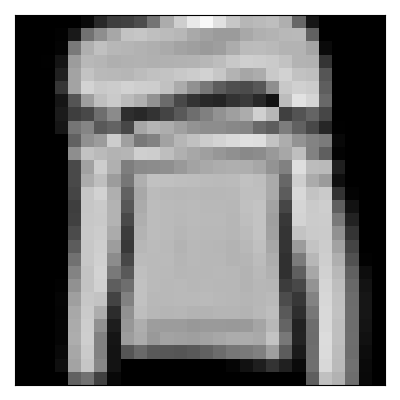}}
& \parbox[c]{1.3cm}{\includegraphics[width=1.3cm]{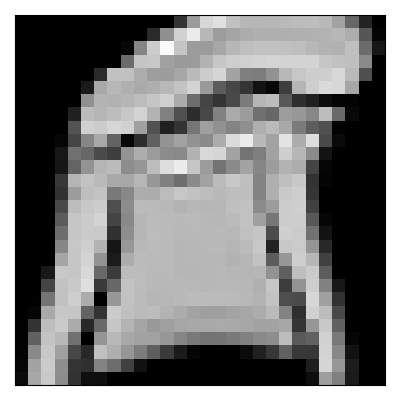}}    \\ \hline
RandomPerspective(distortion\_scale = 0.25 * scale)                            
& \parbox[c]{1.3cm}{\includegraphics[width=1.3cm]{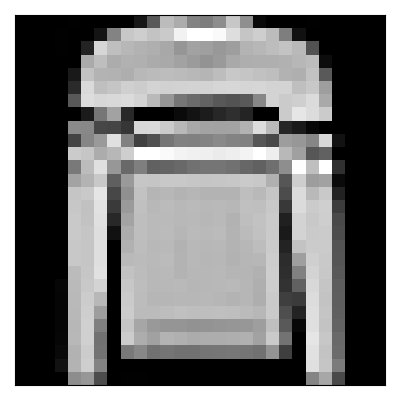}}
& \parbox[c]{1.3cm}{\includegraphics[width=1.3cm]{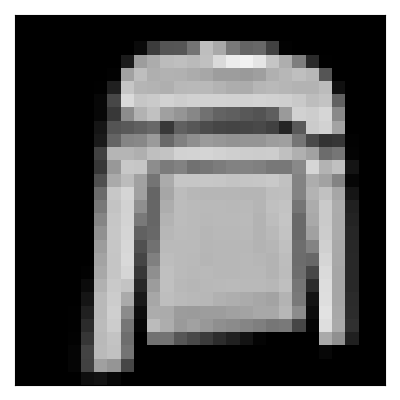}}
& \parbox[c]{1.3cm}{\includegraphics[width=1.3cm]{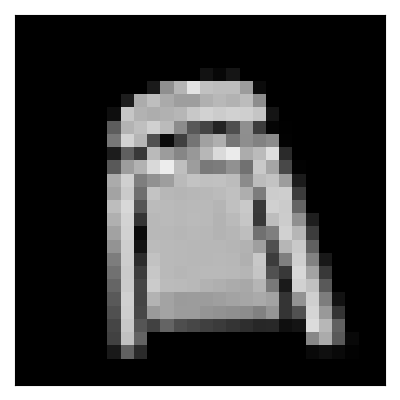}}           \\ \hline
RandomRotation(rotation = 0.60 * scale) 
& \parbox[c]{1.3cm}{\includegraphics[width=1.3cm]{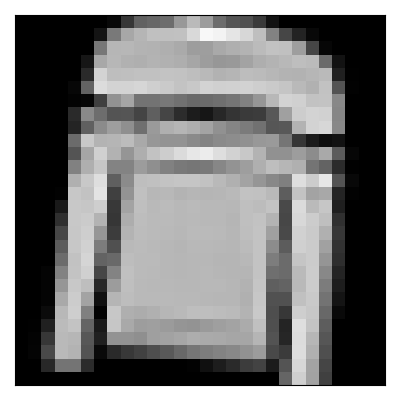}}
& \parbox[c]{1.3cm}{\includegraphics[width=1.3cm]{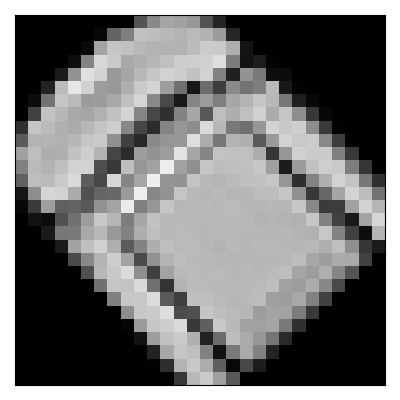}}
& \parbox[c]{1.3cm}{\includegraphics[width=1.3cm]{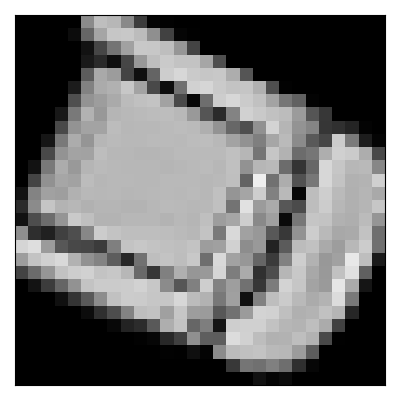}}           \\ \hline
\textbf{Image Modifications by the Authors}
& \textbf{scale=0.2} & \textbf{scale=1.0} & \textbf{scale=2.0} \\ \hline
Obstruction(scale, base\_fraction = 0.40)
& \parbox[c]{1.3cm}{\includegraphics[width=1.3cm]{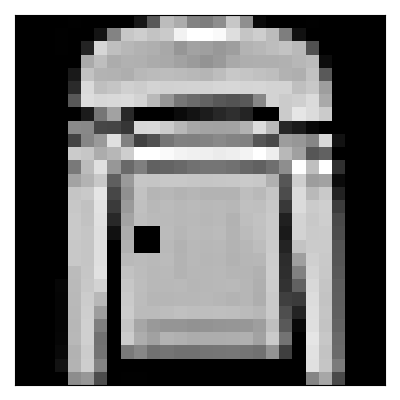}}
& \parbox[c]{1.3cm}{\includegraphics[width=1.3cm]{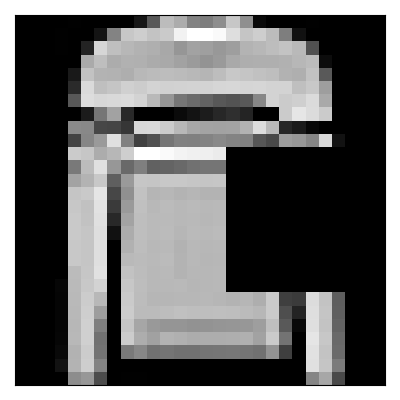}}
& \parbox[c]{1.3cm}{\includegraphics[width=1.3cm]{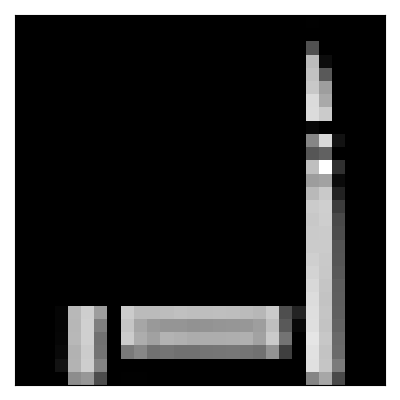}}           \\ \hline
\end{tabular}
}
\centering
    \caption{
    \label{tab:mod_descriptions}
    \textbf{Image modification implementations and examples.}
    }
\end{table*}

\end{document}